%% file: main.tex
\documentclass[journal]{IEEEtran}
\usepackage[ruled,vlined,linesnumbered,longend]{algorithm2e}
\usepackage[utf8]{inputenc}
\usepackage{graphicx}
\usepackage{xcolor}
\usepackage{amsmath,amssymb,amsfonts}
\usepackage{booktabs}  
\usepackage{multirow}
\usepackage{makecell}
\usepackage{enumitem}
\usepackage{etoolbox}
\usepackage{adjustbox}
\usepackage[colorlinks=false, pdfborder={0 0 0}]{hyperref}

\ifCLASSINFOpdf
\else
\fi

\let\origcite\cite
\renewcommand{\cite}[1]{\ifstrempty{#1}{}{\origcite{#1}}}

\begin{document}

\title{Large-Small Model Collaboration for Farmland Semantic Change Detection}

\author{Xinjia~Li$^{\dagger}$,~%
        Rui~Wang$^{\dagger}$,~%
        Qiurong~Peng,~%
        Lingfei~Ye,~%
        Dengrong~Zhang,~%
        and~Haoyu~Zhang$^{*}$%
\thanks{$^{\dagger}$These authors contributed equally to this work.}%
\thanks{$^{*}$Corresponding author: Haoyu Zhang (e-mail: \url{Haoyu.Zhang@hznu.edu.cn}).}%
\thanks{X.~Li, R.~Wang, Q.~Peng, L.~Ye, D.~Zhang, and H.~Zhang are with the College of Information Science and Technology, Hangzhou Normal University, Hangzhou, China.}%
}

\markboth{Journal of \LaTeX\ Class Files,~Vol.~14, No.~8, August~2015}
{Shell \MakeLowercase{\textit{et al.}}: Bare Demo of IEEEtran.cls for IEEE Journals}

\maketitle

\begin{abstract}
Farmland Semantic Change Detection (SCD) is essential for cultivated land protection, yet existing benchmarks and models remain insufficient for fine-grained farmland conversion monitoring. Current datasets often lack dedicated ``from-to'' annotations, while visual change detection models are easily disturbed by phenology-induced pseudo-changes caused by crop rotation, seasonal variation, and illumination differences. To address these challenges, we construct HZNU-FCD, a large-scale fine-grained farmland SCD benchmark with a unified five-class farmland-to-non-farmland annotation protocol. It contains 4,588 bitemporal image pairs with pixel-level labels for practical farmland protection. Based on this benchmark, we propose a large-small collaborative SCD framework that integrates a task-driven small visual model with a frozen large vision-language model. The small model, Fine-grained Difference-aware Mamba (FD-Mamba), learns dense change representations for boundary preservation and small-region localization. The large-model pathway, Cross-modal Logical Arbitration (CMLA), introduces CLIP-based textual priors for prompt-guided semantic arbitration and pseudo-change suppression. To enable effective collaboration, we design a hard-region co-training strategy that supervises the CMLA semantic score map only on low-confidence pixels. Experiments show that our method achieves 97.63\% F1, 96.32\% IoU, and 96.35\% $\mathrm{SCD\_IoU}_{\mathrm{mean}}$ on HZNU-FCD with only 6.65M trainable parameters. Compared with the multimodal ChangeCLIP-ViT, which leverages vision-language information for change detection, our method improves F1 by 10.19 percentage points on HZNU-FCD. It also achieves 91.43\% F1 and 84.21\% IoU on LEVIR-CD, and 93.85\% F1 and 88.41\% IoU on WHU-CD, demonstrating strong robustness and generalization.
The code is available at \url{https://github.com/Lovelymili/FD-Mamba}.
\end{abstract}


\IEEEpeerreviewmaketitle

\input{figure.tex}
\input{table.tex}

\section{Introduction}

\IEEEPARstart{F}{armland} is a fundamental resource for food production and ecological security. Its quantity and spatial distribution are closely related to global food security and regional ecological stability~\cite{Lun06,Sun24b}. In recent years, rapid urbanization and intensive land development have accelerated the conversion of cultivated land into construction land, transportation facilities, and other non-agricultural land-cover types. Therefore, continuous, accurate, and large-scale monitoring of farmland conversion has become an essential technical basis for land resource management and farmland protection. With the rapid development of high-resolution remote sensing imagery, large-area farmland monitoring has become increasingly feasible. However, most existing Binary Change Detection (BCD) methods can only determine whether land-cover changes occur, but cannot identify the semantic category after conversion~\cite{Dau19,Yan22a}. In practical farmland supervision, it is more important to know whether cultivated land has been converted into buildings, roads, bareland, vegetation, or water. This requires SCD methods that provide fine-grained ``from-to'' conversion information, thereby supporting precise farmland protection and land law enforcement~\cite{Sun24b}.

Despite the growing demand for farmland SCD, existing datasets are still insufficient to support fine-grained farmland protection. Most widely used benchmarks, such as LEVIR-CD~\cite{chen2020levir}, SYSU-CD~\cite{Shi21}, and WHU-CD~\cite{ji2018whu}, are mainly designed for urban expansion, building change detection, or general land-cover monitoring. They usually provide binary labels or coarse semantic categories, and therefore cannot explicitly describe farmland-to-non-farmland ``from-to'' conversion relationships. Recent farmland-oriented datasets have made useful progress, but they still have limitations. For example, the Jilin-1 Cup 2024 Second Track dataset~\cite{JL1Cup2024} provides farmland-related SCD labels, but it relies on a single satellite imaging source, which limits its ability to represent cross-source variations in real deployment. Hi-CNA~\cite{Sun24b} is also constructed from single-source satellite imagery, and its coarse category system merges buildings and roads into one ``construction land'' class. This is insufficient for land law enforcement, where different conversion types must be distinguished. Moreover, farmland scenes contain severe phenology-induced pseudo-changes. Crop rotation, seasonal variation, cultivation states, and illumination differences can cause strong color and texture fluctuations between bitemporal images, even when no real land-use conversion occurs. These non-structural variations may be more visually salient than true farmland occupation, causing models trained on general benchmarks to confuse phenological changes with real structural changes. Therefore, a dedicated farmland SCD dataset with unified fine-grained ``from-to'' annotations, dual-source imaging conditions, and careful treatment of pseudo-change interference is urgently needed~\cite{Sui26}.

\FigureFA

Beyond the dataset bottleneck, existing change detection models also face clear limitations when applied to farmland SCD. Early deep learning-based methods mainly relied on convolutional neural networks (CNNs), which are effective in extracting local textures and spatial details from bitemporal images~\cite{Dau18b,Zha17,Zha20b,Fan21,Shi21}. However, their limited receptive fields make it difficult to capture long-range dependencies and global contextual information in high-resolution scenes. Transformer-based methods improve global modeling through self-attention, but their quadratic computational complexity introduces heavy memory and computation costs~\cite{Che21,Ban22,Zha22f,Zha23j,Lei24}. In addition, patch partitioning may disrupt fine spatial structures, especially the continuous boundaries of regular farmland parcels. Recently, Mamba-based models have provided a promising alternative by enabling global context modeling with linear complexity~\cite{Che24h,Zha24l,Zha24k,Don24c}. Nevertheless, their selective scanning mechanism usually flattens two-dimensional feature maps into one-dimensional sequences, which may weaken local geometric connectivity. For farmland scenes with elongated, regular, and densely distributed parcel boundaries, this can lead to fragmented edges and loss of structural information. Moreover, most existing visual models mainly rely on numerical feature differences and lack high-level semantic priors, making them prone to confusing phenology-induced pseudo-changes with real farmland conversion.

Recently, large-small model collaboration has emerged as a promising paradigm for combining the complementary strengths of foundation models and task-specific lightweight models~\cite{chen2025multi}. Large models usually provide rich semantic knowledge and strong representation ability, but they require substantial computational resources and are not always suitable for direct deployment or high-resolution dense prediction. In contrast, small models are efficient and task-adaptive, making them more suitable for pixel-level localization, but their semantic reasoning ability is often limited. Therefore, existing collaborative frameworks connect large and small models through feature alignment, co-training, knowledge transfer, or adaptive prediction fusion, allowing the two pathways to complement each other instead of working independently~\cite{liu2024cotuning,lu2024collaborative,wang2026collaborative}. For example, recent studies integrate off-the-shelf large models with task-driven small models to balance semantic representation and task-specific inference, while tiny-large vision-language frameworks show that mutual learning can improve both efficiency and performance~\cite{wang2022multimodal,chen2024data}. These works demonstrate the potential of large-small collaboration. However, most of them are designed for classification, medical diagnosis, or general vision-language tasks. They are not directly tailored to pixel-level remote sensing SCD, where fine boundary localization, pseudo-change suppression, and semantic ``from-to'' reasoning must be jointly considered.

For farmland SCD, directly introducing large vision-language models is a natural but suboptimal solution. Although such models contain rich semantic priors learned from large-scale image-text data, fully fine-tuning or heavily relying on them is computationally expensive. They may also fail to recover fine parcel boundaries and small changed regions. Meanwhile, lightweight visual models are more efficient and better suited for dense change localization, but they mainly operate within a closed visual feature space. As a result, they may confuse phenology-induced pseudo-changes with real land-use conversion. 

To address the above challenges, we study farmland SCD from both the data and model perspectives. First, we construct HZNU-FCD, a large-scale fine-grained farmland SCD benchmark dedicated to cultivated land protection. To our knowledge, it is the first dataset that integrates self-collected high-resolution UAV imagery and Jilin-1 commercial satellite imagery under a unified five-class farmland-to-non-farmland annotation protocol. HZNU-FCD contains 4,588 bitemporal image pairs with pixel-level labels for practical farmland protection, enabling the evaluation of models under realistic farmland supervision requirements. Second, we propose a large-small collaborative framework for farmland SCD. The task-driven small visual model, Fine-grained Difference-aware Mamba (FD-Mamba), is designed to learn dense visual change representations and preserve fine farmland structures. The large-model branch, Cross-modal Logical Arbitration (CMLA), introduces semantic priors from a frozen vision-language model to verify whether visual changes are logically consistent with real land-cover conversion. Finally, a hard-region co-training strategy is developed to align the two branches by focusing cross-modal supervision on low-confidence pixels, such as ambiguous boundaries, small changed regions, and phenology-induced pseudo-changes.

The main contributions of this paper are summarized as follows:
\begin{itemize}
    \item We construct HZNU-FCD, a large-scale fine-grained SCD dataset dedicated to farmland protection. To our knowledge, it is the first dataset that integrates UAV and satellite imagery under a unified five-class farmland-to-non-farmland ``from-to'' annotation protocol.

    \item We propose a large-small collaborative SCD framework that combines the dense localization ability of a lightweight visual model with the semantic reasoning ability of a frozen vision-language model. FD-Mamba provides fine-grained visual change evidence, while CMLA performs cross-modal logical arbitration to suppress pseudo-changes and enhance real conversion regions.

    \item We design a hard-region co-training alignment strategy that supervises the CMLA semantic score map only on low-confidence pixels from the main visual branch. This enables complementary learning between visual perception and semantic arbitration, improving robustness under complex farmland disturbances.
\end{itemize}

\section{Related Work}

SCD has become an important research topic in remote sensing because it aims to identify not only where changes occur, but also what semantic transitions have taken place. In recent years, deep learning has advanced SCD by improving feature representation, spatio-temporal interaction, and dense prediction accuracy. Existing methods include CNN-based, Transformer-based, Mamba-based, and vision-language-based architectures. Meanwhile, progress in SCD also relies heavily on high-quality benchmark datasets, which provide standardized evaluation protocols and support fair comparison. However, most existing datasets and methods are designed for urban expansion, building change detection, or general land-cover monitoring, and are not specifically tailored to farmland protection. Therefore, this section reviews related work from two aspects: change detection benchmarks and deep learning-based SCD methods.

\subsection{BCD and SCD Benchmarks}

Change detection datasets have evolved from BCD benchmarks to SCD benchmarks. Early BCD datasets provide pixel-level annotations of ``changed'' and ``unchanged'' regions, establishing fundamental benchmarks for change localization~\cite{Dau18b,chen2020levir}. Representative datasets include LEVIR-CD~\cite{chen2020levir} and WHU-CD~\cite{ji2018whu}, which mainly focus on urban building expansion in high-resolution remote sensing images. SYSU-CD~\cite{Shi21} further extends the scene coverage to road construction, vegetation changes, and other land-cover variations. However, binary labels cannot satisfy practical land supervision, where the specific semantic type after conversion must be identified.

To support SCD, several datasets have been constructed to describe semantic transition relationships between land-cover categories. SECOND~\cite{Yan22a} provides both land-cover labels and change information in bitemporal images, enabling models to recognize semantic transitions. HRSCD~\cite{Dau19} further offers large-scale high-resolution images with multi-class semantic annotations. These datasets promote the transition of change detection from coarse binary localization to fine-grained semantic analysis.

Despite their contributions, existing SCD datasets are difficult to directly apply to farmland non-agriculturalization monitoring. General SCD benchmarks, such as SECOND~\cite{Yan22a} and HRSCD~\cite{Dau19}, mainly cover broad land-cover transitions in urban or mixed scenes, and do not explicitly define farmland-oriented ``from-to'' relationships. Recent farmland-specific datasets have made useful progress, but they still have limitations. The Jilin-1 Cup 2024 Second Track dataset~\cite{JL1Cup2024} provides semantic annotations in a farmland-related category space, but it relies on a single satellite imaging source, leaving cross-sensor generalization insufficiently explored. Hi-CNA~\cite{Sun24b} constructs a cropland non-agriculturalization benchmark from GF-2 satellite imagery, but its coarse category system merges buildings and roads into a single class, which is insufficient for land law enforcement.

More recently, Sui \emph{et al.}~\cite{Sui26} constructed a high-resolution pseudo-change benchmark for cropland non-agriculturalization and showed that phenology-driven spectral fluctuations are a major source of false alarms in farmland change detection. Wan \emph{et al.}~\cite{Wan26} further introduced agricultural information-guided priors to filter non-structural spectral variations during farmland change feature extraction. These studies highlight the importance of farmland-specific datasets and pseudo-change suppression, but a large-scale fine-grained farmland SCD benchmark with unified ``from-to'' annotations is still lacking.

\subsection{DL-based SCD}

Deep learning has become the dominant technique for SCD in high-resolution remote sensing images due to its strong feature representation ability. Different from BCD, which only identifies changed regions, SCD further requires recognizing the semantic transition of each changed area, i.e., ``from which land-cover type to which land-cover type''~\cite{Yan22a,Dau19}. Early SCD methods mainly relied on CNNs. For example, HRSCD~\cite{Dau19} established a multi-task learning paradigm by coupling land-cover classification with binary change localization. Later methods, such as SNUNet-CD~\cite{Fan21} and Bi-SRNet~\cite{Din22}, enhanced bitemporal feature interaction through dense skip connections, attention mechanisms, and spatio-temporal reasoning. However, CNN-based methods are limited by local receptive fields and often struggle to capture long-range dependencies in large-scale high-resolution scenes.

To improve global context modeling, Transformer-based methods have been widely introduced into change detection. BiT~\cite{Che21} converts image features into semantic tokens to model global spatio-temporal relationships, while ChangeFormer~\cite{Ban22} adopts a hierarchical encoder-decoder structure for multi-scale representation learning. Although self-attention improves global feature alignment, its quadratic complexity brings heavy computation and memory costs for high-resolution imagery~\cite{Zha22f,Lei24}. In addition, patch partitioning may break fine spatial continuity, which is harmful to narrow and regular farmland parcel boundaries.

Recently, state space models (SSMs), especially Mamba, have provided a promising alternative by modeling long-range dependencies with linear complexity~\cite{Gu23}. Representative methods, such as ChangeMamba~\cite{Che24h}, CDMamba~\cite{Zha24k}, and Mamba-FCS~\cite{wijenayake2026mamba}, explore efficient spatio-temporal interaction, global-local collaboration, and boundary-aware representation. More recent studies further improve Mamba-based change detection by strengthening local geometric modeling. For example, Li \emph{et al.}~\cite{Li26} combine hybrid attention mechanisms with Mamba to enhance local connectivity, while Fang \emph{et al.}~\cite{Fang26} design a local-global encoder with a mask Mamba decoder to improve boundary fidelity in high-resolution change detection.

Meanwhile, multimodal information has gradually attracted attention for SCD. Graph reasoning, vision-language priors, and visual foundation models have been introduced to improve semantic discrimination beyond purely visual difference modeling. HGINet~\cite{HGINet} models hierarchical semantic graph interactions, while CdSC~\cite{CdSC} performs semantic-consistency-constrained difference modeling. ChangeCLIP~\cite{Shi24a} and Semantic-CD~\cite{SemanticCD} introduce CLIP-based textual priors for change understanding. VLCD~\cite{qiu2024novel} explores visual-language foundation models by combining prompt learning, remote sensing feature fusion, and pixel-level change feature computation. VFCCD~\cite{wu2025remote} further incorporates visual foundation model features as semantic constraints to enhance change-region discrimination. In addition, Jia \emph{et al.}~\cite{Jia26} introduce reasoning-based text supervision for language-guided change detection, Cao \emph{et al.}~\cite{Cao26} extend CLIP-based alignment to farmland scenarios under weak supervision, and Dong \emph{et al.}~\cite{Dong26} show that decoupling content and style representations is useful for suppressing non-structural responses caused by illumination and seasonal variation.

Although these methods have achieved promising progress, they are still insufficient for farmland SCD. Farmland scenes contain elongated parcel boundaries, sparse changed regions, subtle inter-class differences, and severe phenology-induced pseudo-changes. Existing visual models may damage local geometric continuity or confuse seasonal fluctuations with real land-use conversion, while current multimodal methods often rely on heavy pretrained backbones or are not tailored to fine-grained farmland ``from-to'' reasoning. These limitations motivate the proposed large-small collaborative framework for robust farmland SCD.

\section{HZNU-FCD Benchmark Construction}

\subsection{Motivation}

Constructing a farmland SCD dataset is challenging due to sparse changes, strong pseudo-changes, and subtle inter-class differences. Farmland non-agriculturalization usually occupies only a small portion of large agricultural scenes, while changed objects such as field roads and parcel boundaries are often narrow and elongated, leading to severe class imbalance. In addition, crop growth, crop rotation, cultivation states, and seasonal illumination can cause significant color and texture variations without real land-use conversion, resulting in phenology-induced pseudo-changes. Categories such as buildings, bareland, and roads may also show similar spectral and texture patterns in high-resolution images, which further increases annotation difficulty. These factors require careful annotation supported by strong scene understanding and multi-source interpretation. Recent work~\cite{Yua26} also confirms that elongated parcel edges are among the most error-prone regions in high-resolution cropland change detection, further motivating edge-sensitive annotation and model design.

\subsection{Data Sources}

HZNU-FCD adopts a dual-source construction strategy to balance fine spatial details and cross-domain generalization. It contains 1,283 pairs of self-collected high-resolution UAV images, together with Jilin-1 commercial satellite image pairs. These data cover diverse geographic regions, sensor types, illumination conditions, and texture scales. After integration, HZNU-FCD includes 4,588 bitemporal image pairs, all cropped into $256 \times 256$ patches. The dataset is randomly split into training, validation, and testing sets with 3,501, 545, and 542 image pairs, respectively.

Table~\ref{tab:dataset_comparison} compares HZNU-FCD with existing farmland-related change detection datasets. The main advantage of HZNU-FCD lies in its dual-source design, which combines low-altitude UAV imagery and orbital satellite imagery to support cross-sensor evaluation. Compared with Hi-CNA~\cite{Sun24b}, HZNU-FCD provides finer-grained five-class annotations and explicitly distinguishes different farmland-to-non-farmland conversion types, such as buildings and roads.

\TabDatasetComparison

\subsection{Semantic Annotation}

HZNU-FCD is designed for practical farmland protection. Unlike coarse binary datasets, it explicitly describes farmland-to-non-farmland semantic transitions by taking the farmland area at time $T_1$ as the reference. Only pixels converted from farmland to other land-cover types are labeled as semantic changes. The dataset contains six pixel-level classes: \begin{itemize}
    \item \textbf{No Change}: farmland remains farmland in both temporal images, and no essential land-use conversion occurs.
    
    \item \textbf{Farmland to Road}: farmland is converted into roads, hardened surfaces, or other transportation-related land.
    
    \item \textbf{Farmland to Building}: farmland is converted into buildings, structures, industrial facilities, or storage land.
    
    \item \textbf{Farmland to Bareland}: farmland becomes bare soil, abandoned land, or land in an early stage of ecological succession.
    
    \item \textbf{Farmland to Vegetation}: farmland is converted into woodland, grassland, or other ecological vegetation land.
    
    \item \textbf{Farmland to Water}: farmland is converted into ponds, aquaculture water surfaces, or water conservancy facilities.
\end{itemize}

To ensure reliable annotations across UAV and satellite imagery, we build a standardized construction workflow with the support of our self-developed Farmland Semantic Change Annotation Platform. The workflow includes candidate region selection, bitemporal image pairing, raw image inspection, label unification, pixel-level annotation, and quality verification. Samples with severe registration errors, occlusions, or incomplete coverage are removed. The original images are first annotated as $512 \times 512$ patches to preserve parcel-level context, and are then cropped into $256 \times 256$ patches for model training and evaluation. Multi-source labels are unified into the proposed six-class semantic system, and isolated change regions smaller than 100 pixels are filtered out to reduce annotation noise.

A three-stage quality control procedure is further adopted, including initial annotation, review, and sampling inspection. For ambiguous samples affected by phenology or complex semantics, historical records and field investigation information are used for cross-validation. This process improves annotation consistency and reduces pseudo-change interference.

\FigureFOne

\section{Large-Small Collaborative Framework}

\FigureFTwo

\subsection{Overview}

To address pseudo-change interference and fine-grained semantic requirements in farmland SCD, we formulate the proposed method as a large-small collaborative framework, as shown in Fig.~\ref{fig:overall}. The framework consists of two complementary components: a task-driven small visual model and a large-model-driven semantic arbitration module. The small model learns efficient dense visual change representations, while the large model provides external semantic priors for logical verification. Through residual semantic gating and hard-region co-training alignment, the two components are optimized collaboratively rather than used as independent predictors.

The small visual model is termed FD-Mamba. It is a lightweight Mamba-based dense prediction model composed of an SRCM-Mamba encoder and a Fine-Grained Difference-Aware Mamba (FGDA-Mamba) decoder. The SRCM-Mamba blocks extract hierarchical structural and contextual features from bitemporal images, while the FGDA-Mamba blocks refine these features into fine-grained change-aware representations. This design enables FD-Mamba to preserve elongated farmland parcel boundaries, recover small changed regions, and suppress low-level visual noise caused by texture and illumination variations.

However, FD-Mamba still mainly operates in the visual feature space. In farmland scenes, non-semantic factors such as shadows, seasonal variation, crop rotation, illumination changes, cloud contamination, sensor noise, and slight misalignment may produce strong visual responses that do not indicate real land-use conversion. Therefore, relying only on the small visual model may still lead to false alarms, especially in phenology-sensitive regions.

To compensate for this limitation, we introduce the CMLA module. CMLA uses a frozen CLIP~\cite{Rad21} text encoder as an external semantic prior provider. Instead of fine-tuning the large model or replacing the visual backbone with a heavy vision-language model, CMLA only uses CLIP to encode task-related textual prompts. The \textit{Brief Prompt} defines the semantic categories of the detection task, such as ``no change'', ``significant land cover change'', or fine-grained ``from-to'' descriptions such as ``farmland change to building'' and ``farmland change to road''. The \textit{Input Prompt} describes the scene context and potential nuisance factors, such as ``Satellite image of suburban area. Ignore shadow.'' These prompts provide semantic constraints for judging whether a visual change is logically consistent with real land-cover conversion.

The collaboration between FD-Mamba and CMLA is achieved in two stages. First, CMLA projects the visual features produced by FD-Mamba into the CLIP embedding space and computes pixel-wise similarities with the adapted text prototypes. The resulting semantic score map is converted into a gate to modulate the original visual features in a residual manner. This allows the semantic prior to enhance real conversion regions while preserving the visual representation learned by the small model. Second, the semantic score map is used as an auxiliary supervision signal only on hard pixels selected by the confidence of the main visual branch. This hard-region co-training strategy avoids redundant supervision on easy regions and encourages CMLA to focus on ambiguous boundaries, small changed objects, and pseudo-change areas.

\subsection{FD-Mamba}

In the proposed large-small collaborative framework, FD-Mamba serves as the task-driven small visual model. It is designed to learn dense and fine-grained visual change representations from bitemporal farmland images. Unlike large vision-language models that provide high-level semantic priors, FD-Mamba focuses on pixel-level localization, boundary preservation, and efficient visual feature learning. This is important for farmland SCD, where true land conversion regions are often sparse, geometrically regular, and easily confused with background disturbances.

FD-Mamba is a unified Mamba-based visual architecture composed of SRCM-Mamba blocks \cite{Zha24k} and FGDA-Mamba blocks. The SRCM-Mamba blocks act as feature encoding units, extracting hierarchical structural and contextual representations from bitemporal inputs. These features capture farmland textures, parcel boundaries, field roads, and high-level land-cover context. The FGDA-Mamba blocks then serve as decoding units that transform the encoded bitemporal features into fine-grained difference-aware representations. Instead of relying on simple bitemporal differencing, FGDA-Mamba progressively refines change responses through difference anchoring, multi-scale spatial excavation, attention-based purification, redundancy suppression, and Conv-Mamba reconstruction.

As shown in Fig.~\ref{fig:overall}, the refinement process inside each FGDA-Mamba block is progressive. The Physical Difference Anchor(PDA) first produces an initial difference feature, where building outlines and road structures are visible but still mixed with scattered background noise. The Multi-scale Spatial Difference Excavation(MSDE) aggregates local details through multi-scale grouped convolutions, making change regions more continuous and boundaries clearer. The Dual-Path Difference Purification and Saliency Enhancement(DPSE) module further recalibrates features from both channel and spatial dimensions, enhancing change-related responses and suppressing irrelevant activations. The Dimensionality Reduction and Redundancy Suppression Aggregation(DRSA) module reduces redundant background information after feature fusion and generates a compact representation. Finally, Conv-Mamba performs global context refinement, further suppressing high-frequency pseudo-changes and concentrating responses on real changed regions. Through this design, FD-Mamba combines efficient dense prediction with Mamba-based long-range modeling, providing reliable visual evidence for subsequent CMLA-based cross-modal logical arbitration.

\subsubsection{PDA}

Direct feature differencing is widely used in change detection, but it may amplify phenology-induced variations and weaken stable structures in farmland scenes. To obtain a more reliable difference representation, we introduce the Physical Difference Anchor (PDA), which combines numerical differences with stable structural responses from bitemporal features:

$$
P^i =
\mathrm{Concat}
\left(
|I_a^i - I_b^i|,
\mathrm{Max}(I_a^i,I_b^i)
\right),
$$

where $I_a^i$ and $I_b^i$ denote the bitemporal features at the $i$-th scale. $|\cdot|$ is the element-wise absolute operation, $\mathrm{Max}(\cdot)$ is the element-wise maximum operation, and $\mathrm{Concat}(\cdot)$ denotes channel-wise concatenation. The absolute difference highlights potential change cues and avoids sign ambiguity, while the maximum response preserves salient geometric structures, such as parcel boundaries, field roads, and building contours. Thus, PDA provides a stable input for subsequent difference excavation.
 
\subsubsection{MSDE}

Farmland changes show large scale variations. Narrow field roads and parcel boundaries require fine local perception, while building expansion, bareland formation, and water occupation require broader contextual modeling. To capture these complementary patterns, we design the Multi-scale Spatial Difference Excavation (MSDE) module, which extracts multi-scale spatial differences from the PDA feature through two parallel grouped convolution branches:

$$
M^i =
\mathrm{Conv}_{1 \times 1}
\left(
\mathrm{Concat}
\left(
\mathrm{GConv}_{3 \times 3}(P^i),
\mathrm{DGConv}_{3 \times 3}^{d=2}(P^i)
\right)
\right).
$$

Here, $\mathrm{GConv}_{3 \times 3}(\cdot)$ denotes a standard $3 \times 3$ grouped convolution, and $\mathrm{DGConv}_{3 \times 3}^{d=2}(\cdot)$ denotes a dilated $3 \times 3$ grouped convolution with dilation rate $d=2$. The standard branch captures fine-grained boundaries and small changed regions, while the dilated branch provides a larger receptive field for regional context modeling without increasing the kernel size. The final $1 \times 1$ convolution performs channel projection and feature fusion. In this way, MSDE efficiently enhances multi-scale difference representation for high-resolution farmland imagery.

\subsubsection{DPSE}

The multi-scale difference feature $M^i$ may still contain pseudo-change responses caused by illumination variation, crop phenology, soil moisture, and texture disturbance. To suppress these noisy responses before Mamba-based global modeling, we design the Dual-Path Difference Purification and Saliency Enhancement (DPSE) module. DPSE recalibrates difference features from both channel and spatial dimensions by combining SE attention and CBAM attention:

$$
R^i =
\mathrm{Conv}_{1 \times 1}
\left(
M^i \odot \mathrm{SE}(M^i) \odot \mathrm{CBAM}(M^i)
\right).
$$

Here, $\odot$ denotes element-wise multiplication. $\mathrm{SE}(\cdot)$ selects discriminative change-related channels and suppresses redundant background responses, while $\mathrm{CBAM}(\cdot)$ enhances spatial regions that are more likely to contain real structural changes. The output $R^i$ is the purified and saliency-enhanced difference feature.

\subsubsection{DRSA}

Although the purified difference feature highlights potential change cues, the original bitemporal features still provide useful context for distinguishing spectrally similar categories, such as buildings, bareland, and roads. Therefore, the Dimensionality Reduction and Redundancy Suppression Aggregation (DRSA) module fuses the purified difference feature with the original bitemporal features to retain both change evidence and temporal context.

Direct concatenation, however, increases the channel dimension and introduces redundant background responses. These redundant features may increase the computational cost of subsequent Mamba modeling and weaken the discriminative change cues. To address this issue, DRSA first uses a $1 \times 1$ convolution to compress the fused feature into a unified channel space, and then applies a local convolution for spatial refinement. An SE attention operation is further used to recalibrate the refined feature along the channel dimension. This process preserves useful bitemporal context while suppressing redundant background information, producing a compact change-aware representation for Conv-Mamba reconstruction.

After DRSA, the Conv-Mamba reconstruction unit further refines the compact representation by combining Mamba-based long-range dependency modeling with convolutional local detail perception. This step suppresses residual high-frequency pseudo-changes and concentrates responses on real changed regions. The final output of the FGDA-Mamba decoder is used as the visual change representation and is then passed to CMLA for semantic consistency verification.

\subsection{Cross-modal Logical Arbitration Module}

Although FD-Mamba can generate refined visual change features, it still mainly relies on visual evidence. In farmland scenes, shadows, illumination variation, seasonal changes, crop phenology, and sensor noise may produce strong visual differences that do not indicate real land-cover conversion. To suppress these non-semantic pseudo-change responses, we design the CMLA module. CMLA uses a frozen CLIP text encoder~\cite{Rad21} as an external semantic prior and performs pixel-wise vision-language matching between decoder features and text-defined change categories. The resulting semantic score map is used for residual gating, so that semantically consistent change regions are enhanced while nuisance-induced pseudo-changes are suppressed.

As shown in Fig.~\ref{fig:overall}, CMLA takes three inputs: the decoder output feature $Z$, the Brief Prompt, and the Input Prompt. The Brief Prompt defines the target categories of the task. For BCD, it contains ``no change'' and ``significant land cover change''. For fine-grained SCD, it is extended to dataset-specific ``from-to'' descriptions, such as ``farmland change to bareland'', ``farmland change to building'', ``farmland change to road'', ``farmland change to vegetation'', and ``farmland change to water''. In contrast, the Input Prompt describes the scene context and potential nuisance factors of the input images. It is constructed from geo-scene and geo-nuisance labels generated by Gemini 3-Pro-Preview~\cite{geminiteam2024gemini}.

Specifically, the Input Prompt follows the template ``Satellite image of \{scene\} area. Ignore \{nuisance\_1\}, \{nuisance\_2\}, ...''. The scene label is selected from eight categories: urban, suburban, rural, forest, farmland, water, mixed, and unknown. The nuisance factors include shadow, illumination, season, misalignment, cloud, and sensor noise. A nuisance factor is included only when its confidence is higher than $0.5$. If no obvious nuisance exists, the prompt is simplified as ``Clear conditions.'' Notably, the Input Prompt deliberately contains no change-type information, which avoids label leakage. Its role is to guide the model to ignore specific nuisance factors under a given scene context, rather than to search for a specific change category.

CMLA contains five main operations. First, the fixed Brief Prompt is encoded by the frozen CLIP text encoder to obtain category prototypes:

$$
t_{\mathrm{cat}}
=
\mathrm{Norm}
\left(
\mathcal{E}_{\mathrm{txt}}(\mathcal{P}_{\mathrm{brief}})
\right),
$$

where $\mathcal{P}_{\mathrm{brief}}$ denotes the Brief Prompt set, $\mathcal{E}_{\mathrm{txt}}(\cdot)$ is the frozen CLIP text encoder, and $t_{\mathrm{cat}} \in \mathbb{R}^{N \times d}$ denotes the category prototype matrix. Here, $N$ is the number of semantic categories and $d$ is the CLIP embedding dimension. Since the Brief Prompt is fixed, $t_{\mathrm{cat}}$ can be pre-computed and reused during training.

Second, the Input Prompts in the current batch are encoded by the same frozen CLIP text encoder. Their normalized representations are averaged and passed through a lightweight MLP to generate a shared context offset:

$$
\delta
=
\mathrm{MLP}
\left(
\frac{1}{B}
\sum_{i=1}^{B}
\mathrm{Norm}
\left(
\mathcal{E}_{\mathrm{txt}}
\left(
\mathcal{P}_{\mathrm{input}}^{(i)}
\right)
\right)
\right),
$$

where $\mathcal{P}_{\mathrm{input}}^{(i)}$ denotes the Input Prompt of the $i$-th sample and $B$ is the batch size. The offset $\delta \in \mathbb{R}^{d}$ adapts the category prototypes according to the dominant scene and nuisance characteristics of the batch. The adapted prototypes are obtained by residual correction:

$$
\tilde{t}_{\mathrm{cat}}
=
\mathrm{Norm}
\left(
t_{\mathrm{cat}} + \alpha \delta
\right),
$$

where $\alpha$ is a learnable scaling factor that controls the strength of context adaptation.

Third, the decoder output feature $Z$ is projected into the CLIP embedding space by a $1 \times 1$ convolution and then normalized:

$$
V
=
\mathrm{Norm}
\left(
\mathrm{Conv}_{1 \times 1}(Z)
\right),
$$

where $V$ denotes the pixel-level visual embedding. This projection aligns the visual feature with the text prototype space.

Fourth, pixel-wise vision-language similarity is computed between $V$ and the adapted category prototypes $\tilde{t}_{\mathrm{cat}}$ by an Einsum operation:

$$
S
=
\gamma \cdot
\mathrm{Einsum}
\left(
V,\tilde{t}_{\mathrm{cat}}
\right),
$$

where $\gamma$ is a learnable temperature parameter and $S$ denotes the semantic score map. The score map measures the semantic consistency between each pixel and each text-defined category prototype.

Finally, the semantic score map is converted into a pixel-level gate:

$$
G
=
\mathrm{Sigmoid}
\left(
\mathrm{Conv}_{1 \times 1}(S)
\right).
$$

The gate modulates the original visual feature in a residual manner:

$$
Z_g
=
Z + Z \odot G,
$$

where $\odot$ denotes element-wise multiplication. This residual design preserves the original visual representation while enhancing semantically reliable responses. It also prevents true changes from being over-suppressed when the text branch is imperfect. The gated feature $Z_g$ is then fed into the main classification head to produce the final prediction.

In addition to semantic gating, the score map $S$ is used as an auxiliary supervision signal during training. To avoid redundant supervision with the main visual branch, we apply this auxiliary constraint only to low-confidence pixels selected by the main prediction. This encourages CMLA to focus on ambiguous boundaries, small changed regions, and pseudo-change areas caused by shadows, illumination, or seasonal variation.

\subsection{Co-training Alignment}

The CMLA module produces a semantic score map that is used for both residual gating and auxiliary supervision. However, applying full-image supervision to both the main visual branch and the CMLA branch may cause redundant optimization, especially for easy pixels that are already correctly classified. To promote complementary learning, we design a confidence-based hard-region auxiliary training strategy.

The main branch is supervised by a standard segmentation loss that combines cross-entropy loss and Dice loss:

$$
\mathcal{L}_{\mathrm{main}}
=
0.5\mathcal{L}_{\mathrm{CE}}(\hat{Y},Y)
+
0.5\mathcal{L}_{\mathrm{Dice}}(\hat{Y},Y),
$$

where $\hat{Y}$ denotes the final prediction logits and $Y$ denotes the ground-truth semantic change map. Cross-entropy provides pixel-level semantic supervision, while Dice loss alleviates class imbalance caused by sparse farmland conversion regions. Invalid pixels with label value $255$ are ignored.

To guide CMLA toward uncertain regions, we construct a hard-region mask from the confidence of the main prediction. Let $p=\mathrm{Softmax}(\hat{Y})$ be the main-branch prediction probability. Pixels whose maximum class probability is lower than the threshold $\tau$ are selected as hard samples:

$$
M_{i,j}^{\mathrm{hard}}
=
\mathbf{1}
\left(
\max_k p_{i,j,k} < \tau
\right),
$$

where $\tau$ is set to $0.80$ by default. These pixels usually correspond to ambiguous parcel boundaries, small changed objects, spectrally similar categories, or pseudo-change regions caused by shadows, illumination, and seasonal variation.

Given the semantic score map $S$ generated by CMLA, the auxiliary loss is applied only to the selected hard regions:

$$
\mathcal{L}_{\mathrm{aux}}
=
\frac{
\sum_{i,j}
M_{i,j}^{\mathrm{hard}}
\mathcal{L}_{\mathrm{CE}}(S_{i,j},Y_{i,j})
}{
\sum_{i,j}M_{i,j}^{\mathrm{hard}}+\epsilon
},
$$

where $\epsilon$ avoids division by zero. This design prevents CMLA from repeatedly learning easy background pixels and encourages it to focus on semantically ambiguous regions where textual priors are more useful.

The final objective is defined as:

$$
\mathcal{L}_{\mathrm{total}}
=
\mathcal{L}_{\mathrm{main}}
+
\lambda \mathcal{L}_{\mathrm{aux}},
$$

where $\lambda$ is set to $0.4$ in our experiments. Through this co-training alignment strategy, the visual branch provides reliable dense predictions, while CMLA offers complementary semantic guidance on hard pixels, improving robustness under phenological pseudo-changes and complex farmland boundaries.

\section{Experimental Settings}

\subsection{Benchmark Datasets}

We evaluate the proposed framework on two change detection tasks: SCD and BCD. For SCD, we use the proposed HZNU-FCD dataset to assess performance in practical farmland protection scenarios. For BCD, we conduct experiments on two widely used building change detection benchmarks, LEVIR-CD and WHU-CD, to verify the generalization ability of the proposed method.

\textbf{HZNU-FCD.}
As introduced in Section~3, HZNU-FCD is the core SCD benchmark in this work. It contains 4,588 high-resolution bitemporal image pairs with a size of $256 \times 256$ pixels. The dataset covers diverse geographic regions, imaging conditions, and phenology-induced disturbances. It provides pixel-level fine-grained ``from-to'' annotations with six categories, including no change and five farmland-to-non-farmland conversion types. We use 3,501 image pairs for training, 545 for validation, and 542 for testing.

\textbf{LEVIR-CD.}
LEVIR-CD~\cite{chen2020levir} is a representative high-resolution BCD benchmark for building change detection. It contains 637 pairs of Google Earth images with a spatial resolution of 0.5 m and an original size of $1024 \times 1024$ pixels. Following the common protocol, we crop the images into non-overlapping $256 \times 256$ patches and use the official 7:1:2 split for training, validation, and testing.

\textbf{WHU-CD.}
WHU-CD~\cite{ji2018whu} is another widely used BCD benchmark for building change detection. It consists of aerial images acquired in 2012 and 2016 over Christchurch, New Zealand, with a spatial resolution of 0.2 m and an original size of $32507 \times 15354$ pixels. We crop the images into non-overlapping $256 \times 256$ patches and follow the standard split for training and evaluation.

\subsection{Baselines}

\subsubsection{Baseline Models}

We compare the proposed framework with representative change detection methods from four categories: CNN-based, Transformer-based, Mamba-based, and multimodal methods. The CNN-based baselines include MCTNet~\cite{MCTNet} and SNUNet~\cite{Fan21}. The Transformer-based baselines include BIT~\cite{Che21}, ChangeFormer~\cite{Ban22}, and SCanNet~\cite{SCanNet}. The Mamba-based baselines include MambaFCS~\cite{MambaFCS}, ChangeMamba~\cite{Che24h}, and CDMamba~\cite{Zha24k}. In addition, ChangeCLIP-RN50 and ChangeCLIP-ViT~\cite{Shi24a} are selected as multimodal baselines to evaluate the effect of vision-language priors.

\subsubsection{Evaluation Metrics}

For BCD benchmarks, including LEVIR-CD and WHU-CD, we adopt Precision, Recall, F1-score, Intersection over Union (IoU), and Overall Accuracy (OA), following standard protocols~\cite{chen2020levir,ji2018whu}. For the farmland SCD benchmark HZNU-FCD, we further use semantic change F1-score ($F_{\mathrm{scd}}$), semantic kappa coefficient (SeK), and mean semantic change IoU ($\mathrm{SCD\_IoU}_{\mathrm{mean}}$), following~\cite{Yan22a}. These metrics jointly measure change localization and fine-grained semantic recognition, making them suitable for evaluating farmland ``from-to'' conversion.

\subsection{Implementation Details}

All input image pairs are resized or cropped to $256 \times 256$ pixels for training and evaluation. The proposed model is optimized with AdamW using an initial learning rate of $1 \times 10^{-4}$. A cosine annealing schedule is adopted for learning rate adjustment, and all models are trained for 300 epochs.

The training objective follows the co-training alignment strategy described in Section~4.4. The main branch is supervised by a combination of cross-entropy loss and Dice loss to jointly optimize pixel-level classification and region-level change consistency. In addition, a hard-region auxiliary cross-entropy loss is applied to the CMLA semantic score map based on the confidence of the main prediction. This auxiliary loss encourages CMLA to focus on difficult pixels, such as ambiguous boundaries, small changed regions, and phenology-induced pseudo-change areas. The auxiliary loss weight is set to $\lambda=0.4$, and the hard-region confidence threshold is set to $\tau=0.80$.

\section{Experimental Results}

In this section, we report experimental results on the farmland-oriented HZNU-FCD dataset and two BCD benchmarks, LEVIR-CD and WHU-CD. We first compare the proposed method with representative baselines quantitatively and qualitatively. Then, we conduct ablation studies to evaluate the contributions of FGDA-Mamba, CMLA, and the auxiliary loss settings.

\subsection{Results on HZNU-FCD}

\TabHZNUFCDCompare

Table~\ref{tab:hznu_fcd_compare} reports the quantitative results on the HZNU-FCD test set. The compared methods cover CNN-based, Transformer-based, hybrid, Mamba-based, and vision-language multimodal architectures. HZNU-FCD is more challenging than common binary building benchmarks because it requires models to handle sparse changes, fine-grained semantic transitions, and phenology-induced pseudo-changes. Our method achieves the best F1 of 97.63\%, IoU of 96.32\%, OA of 98.28\%, $F_{\mathrm{scd}}$ of 96.87\%, and $\mathrm{SCD\_IoU}_{\mathrm{mean}}$ of 96.35\%, with only 6.65M trainable parameters. Compared with the best CNN-based baseline, SNUNet, our method improves F1 and IoU by 17.97 and 30.12 percentage points, respectively. Compared with the best Transformer-based baseline, CdSC, it further improves F1 and IoU by 3.79 and 7.93 percentage points. Against the strongest Mamba-based baseline, CDMamba, our method still achieves gains of 0.13, 0.22, and 0.19 percentage points in F1, IoU, and $\mathrm{SCD\_IoU}_{\mathrm{mean}}$, respectively, while using only 55.9\% of its trainable parameters. These results show that the proposed framework achieves a favorable balance between accuracy and efficiency.

\FigureQualHZNU

Fig.~\ref{fig:qual_hznu} presents qualitative results covering all five farmland-to-non-farmland conversion categories. MCTNet shows semantic confusion around road--bareland intersections and often produces broken parcel boundaries. MambaFCS improves global modeling, but still generates fragmented predictions along narrow roads and water boundaries. ChangeCLIP-RN50 and ChangeCLIP-ViT introduce vision-language priors, yet their predictions remain unstable for fine-grained ``from-to'' categories and often contain scattered labels within the same changed region. CDMamba recovers most large changed regions, but may over-extend road boundaries into adjacent farmland and miss small farmland-to-water conversions. In contrast, our method produces more compact and coherent predictions. Road conversions remain continuous, narrow water-related changes are better recovered, and intra-class consistency within changed parcels is improved. These results indicate that FD-Mamba provides reliable structural change representations, while CMLA further enhances semantic discrimination between spectrally similar categories such as bareland and road.

\subsection{Results on LEVIR-CD}

\TabLEVIRCompare

To evaluate the generalization ability of the proposed framework, we conduct experiments on LEVIR-CD, a widely used building change detection benchmark. As shown in Table~\ref{tab:levir_compare}, our method achieves 92.69\% Precision, 90.20\% Recall, 91.43\% F1, and 84.21\% IoU with only 6.65M trainable parameters. Compared with FC-Siam-Conc, the best CNN-based baseline, our method improves F1 and IoU by 1.54 and 2.57 percentage points, respectively. Compared with DARNet, the best Transformer-based baseline, it further improves F1 and IoU by 0.87 and 1.45 percentage points. Against the strong Mamba-based baseline CDMamba, our method improves F1 and IoU by 0.68 and 1.14 percentage points while using only about 55.9\% of its parameters. These results show that the proposed framework achieves a strong balance between accuracy and efficiency.

\FigureQualLEVIR

Fig.~\ref{fig:qual_levir} shows representative qualitative results. BiT and ChangeMamba can detect large building changes, but they often produce fragmented regions or boundary bleeding in dense building areas. CDMamba improves global consistency, yet it may merge adjacent buildings and miss small structures. ChangeCLIP-based methods benefit from vision-language priors, but still generate false positives under strong shadow and illumination variations. In contrast, our method better separates adjacent buildings, preserves sharper boundaries, and suppresses shadow-induced false alarms. This demonstrates the effectiveness of FD-Mamba in fine-grained structural modeling and CMLA in pixel-level semantic verification.

\subsection{Results on WHU-CD}

WHU-CD contains more irregular building boundaries and stronger background interference than LEVIR-CD, making it a stricter benchmark for evaluating model robustness.

\TabWHUCompare

As shown in Table~\ref{tab:whu_compare}, our method achieves the best overall performance, with 94.97\% Precision, 92.75\% Recall, 93.85\% F1, 88.41\% IoU, and 99.52\% OA. Compared with the best CNN-based baseline, FC-EF, our method improves F1 and IoU by 2.49 and 4.31 percentage points, respectively. Compared with the best Transformer-based baseline, PaFormer, it further improves F1 and IoU by 1.56 and 2.72 percentage points. Against the strongest Mamba-based baseline, CDMamba, our method achieves gains of 0.74, 0.09, and 0.15 percentage points in Recall, F1, and IoU, respectively. These results indicate that the proposed framework improves region completeness and boundary recovery while maintaining strong noise suppression in complex building scenes.

\FigureQualWHU

Fig.~\ref{fig:qual_whu} presents representative qualitative results on WHU-CD. ChangeFormer often misses irregular building boundaries and produces fragmented changed regions. ChangeMamba improves global coverage, but still loses thin rooftop structures and generates false positives along illumination boundaries. CDMamba produces cleaner contours, but may under-segment densely packed buildings and slightly over-extend into shadow areas. In contrast, our method provides more complete building footprints, sharper boundaries, and fewer shadow-induced false alarms. This benefits from the structural refinement of FD-Mamba and the semantic verification of CMLA, which jointly improve region completeness and suppress background interference.

\subsection{Effectiveness of the FGDA-Mamba Decoder}

To evaluate the contribution of the proposed FGDA-Mamba decoder, we conduct both component-level and cross-dataset ablation studies. For component analysis, we remove one key module at a time on the HZNU-FCD test set while keeping the remaining modules unchanged. The results are reported in Table~\ref{tab:abl_fgda_components}.

\TabAblFGDAComponents

The results show that MSDE, DPSE, and DRSA all contribute positively to the final performance. Removing DPSE causes the largest degradation, with F1, IoU, and $\mathrm{SCD\_IoU}_{\mathrm{mean}}$ decreasing by 0.78, 2.44, and 1.42 percentage points, respectively. This confirms the importance of DPSE in suppressing phenology- and illumination-induced pseudo-change responses. Removing MSDE also leads to consistent drops in F1 and IoU by 0.27 and 1.47 percentage points, showing that multi-scale spatial difference modeling is important for narrow roads, parcel boundaries, and large-area conversions. DRSA brings a smaller but still meaningful improvement by reducing redundant background responses and improving feature compactness. Overall, MSDE enhances structural perception, DPSE purifies noisy difference responses, and DRSA suppresses redundancy. Their combination enables the full FGDA-Mamba decoder to achieve the best performance.

To further verify its generality, we replace the original decoder of CDMamba with the proposed FGDA-Mamba decoder and evaluate it on LEVIR-CD, WHU-CD, and HZNU-FCD. On LEVIR-CD, FGDA-Mamba achieves 90.99\% F1 and 83.47\% IoU, outperforming the original CDMamba. On WHU-CD, it obtains 92.88\% F1 and 86.71\% IoU, showing stronger region recovery under irregular building boundaries. On HZNU-FCD, the full FGDA-Mamba decoder improves F1, IoU, and $F_{\mathrm{scd}}$ over the baseline by 0.97, 2.78, and 1.24 percentage points, respectively. These results demonstrate that FGDA-Mamba consistently improves change representation across both BCD and SCD tasks, providing a strong visual foundation for the subsequent CMLA module.

\subsection{Effectiveness of CMLA Module}

We further evaluate the contribution of the CMLA module on both BCD and SCD tasks.

\subsubsection{Performance on the BCD Task}

For BCD, CMLA uses the text prototypes ``no change'' and ``significant land cover change'' to perform pixel-level semantic verification on the visual features produced by the FGDA-Mamba decoder. As shown in Table~\ref{tab:abl_CMLA_bcd}, adding CMLA consistently improves performance on both LEVIR-CD and WHU-CD. On LEVIR-CD, F1 and IoU increase from 90.99\% and 83.47\% to 91.74\% and 84.75\%, respectively. On WHU-CD, F1 and IoU improve from 92.88\% and 86.71\% to 94.38\% and 89.35\%, respectively.

The qualitative results in Figs.~\ref{fig:abl_CMLA_bcd_levir} and~\ref{fig:abl_CMLA_bcd_whu} further explain these gains. On LEVIR-CD, CMLA suppresses false positives caused by shadows and illumination changes. On WHU-CD, it helps recover missed building regions with coherent structural changes. This improvement comes from the semantic gating mechanism: nuisance regions usually show low similarity to the ``significant land cover change'' prototype, while real structural changes obtain higher prototype-matching scores. The context adapter further adjusts the prototypes according to batch-level nuisance information, improving the separation between real changes and non-semantic visual disturbances.

\TabAblCMLABCD

\FigureAblCMLABCDLEVIR
\FigureAblCMLABCDWHU

\subsubsection{Performance on the SCD Task}

For SCD, CMLA is more important because the model must identify both change locations and specific ``from-to'' categories. It constructs category-wise text prototypes for each farmland conversion type and uses pixel-level vision-language matching to guide semantic discrimination. As shown in Table~\ref{tab:abl_CMLA_scd}, adding CMLA improves F1 from 97.50\% to 97.53\%, IoU from 96.10\% to 96.19\%, and OA from 98.19\% to 98.28\%. Although $F_{\mathrm{scd}}$ slightly decreases, the improvement in region-level metrics indicates that CMLA mainly refines changed-parcel boundaries and strengthens category-level completeness.

The qualitative results in Fig.~\ref{fig:abl_CMLA_scd} further show that CMLA reduces confusion between spectrally similar classes, such as road and bareland, and suppresses phenology-induced pseudo-changes. This is because the ``from-to'' text prototypes provide semantic constraints beyond visual color and texture. Pseudo-change regions caused by crop rotation or seasonal variation usually show low similarity to all real-conversion prototypes, and are therefore weakened by the semantic gate. The context adapter further adjusts the prototypes according to batch-level nuisance information, improving robustness to seasonal and illumination disturbances. These results demonstrate that CMLA enhances semantic consistency without introducing additional visual parameters.

\TabAblCMLASCD

\FigureAblCMLASCD

\subsection{Hyperparameter Analysis of the Auxiliary Loss}

We analyze the sensitivity of two key hyperparameters in the hard-region auxiliary loss: the confidence threshold $\tau$ and the auxiliary weight $\lambda$. The threshold $\tau$ determines which pixels are selected as hard samples according to the main-branch confidence, while $\lambda$ controls the contribution of CMLA auxiliary supervision. We evaluate six configurations on LEVIR-CD by fixing one parameter and varying the other.

\TabAblTauLambda

As shown in Table~\ref{tab:abl_tau_lambda}, $\tau=0.80$ and $\lambda=0.40$ achieve the best overall balance, with 91.43\% F1, 84.21\% IoU, and 99.14\% OA. A smaller $\lambda$ provides limited semantic guidance, while a larger $\lambda$ may interfere with the optimization of the main visual branch. Similarly, a smaller $\tau$ selects fewer hard pixels and may reduce change coverage, whereas a larger $\tau$ introduces more uncertain pixels and may slightly weaken precision. Therefore, we use $\tau=0.80$ and $\lambda=0.40$ as the default setting in all experiments.

Overall, FD-Mamba and CMLA play complementary roles. FD-Mamba constructs compact and structurally consistent visual change representations, while CMLA provides cross-modal semantic verification to suppress pseudo-changes and enhance category consistency. With hard-region co-training, the proposed framework achieves accurate localization and reliable semantic recognition across both BCD and farmland SCD tasks with only 6.65M trainable parameters.

\section{Conclusion}

In this paper, we present HZNU-FCD, a large-scale fine-grained farmland SCD dataset with unified farmland-to-non-farmland annotations. We further propose a large-small collaborative framework for robust farmland change detection. In this framework, FD-Mamba learns compact and structurally consistent visual change representations, while CMLA introduces CLIP-based semantic priors to suppress pseudo-changes. A hard-region co-training strategy aligns the two branches by focusing semantic supervision on difficult pixels. Experiments on HZNU-FCD, LEVIR-CD, and WHU-CD demonstrate that our method achieves strong accuracy and generalization with only 6.65M trainable parameters. Future work will extend HZNU-FCD to more regions and imaging conditions, and explore open-vocabulary farmland change understanding.

\bibliographystyle{IEEEtran}
\bibliography{references}

\end{document}

%% file: figure.tex
\newcommand{\FigurePhenology}{%
\begin{figure*}[t]
  \centering
  \includegraphics[width=\textwidth]{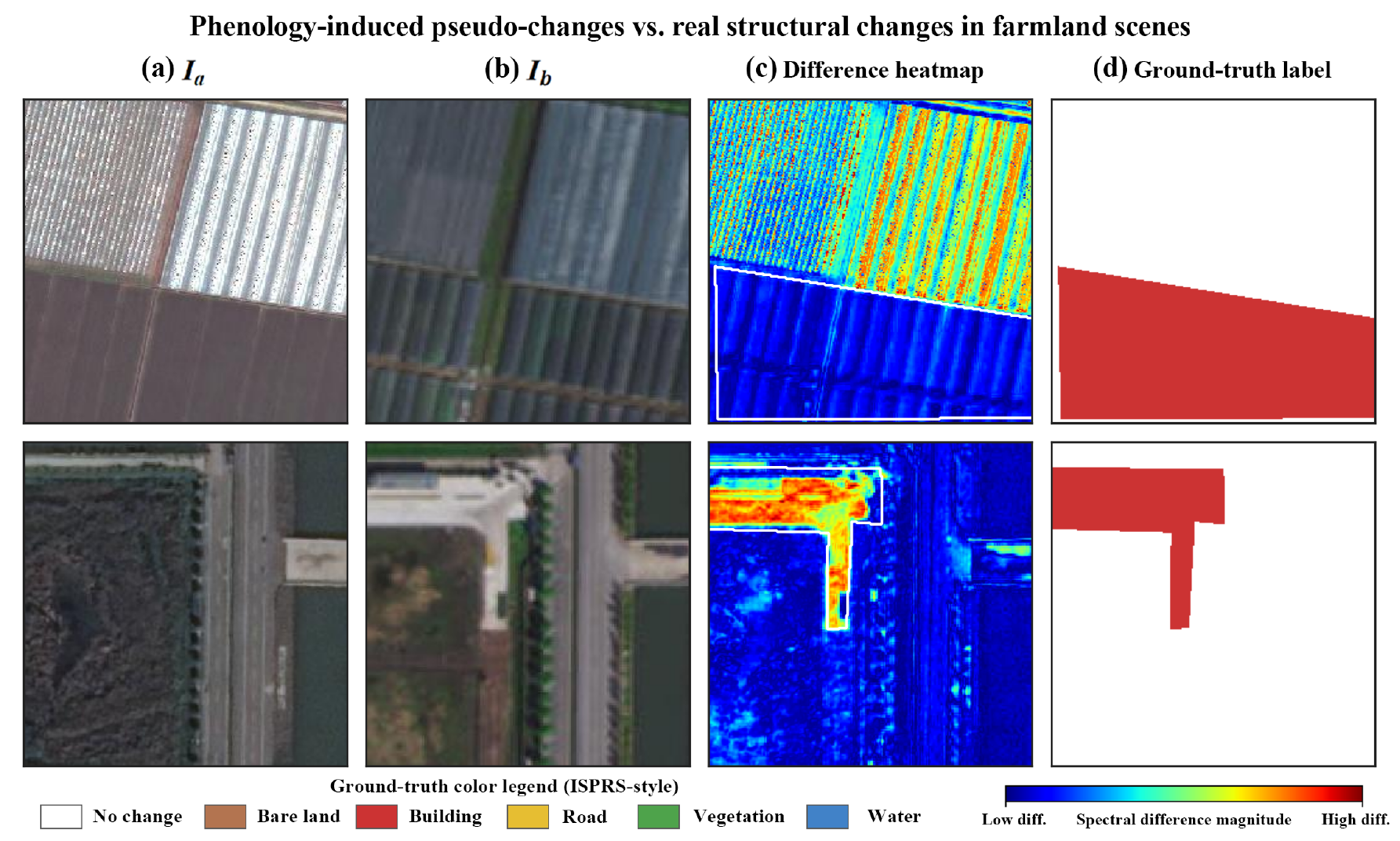}
  \caption{Illustration of phenology-induced pseudo-changes in farmland scenes from the HZNU-FCD dataset. (a)--(b)~Bi-temporal images $T_1$ and $T_2$; (c)~pixel-wise spectral difference heatmap (warm colors indicate large spectral difference; white contour marks the true change boundary); (d)~ground-truth semantic label. \textit{Top row}: The greenhouse strips are land-use-\emph{unchanged}, yet exhibit drastic spectral shifts between $T_1$ and $T_2$ due to seasonal crop rotation and illumination variation, producing high heatmap activations over the unchanged area while the actual structural change occupies only the small region inside the white contour. \textit{Bottom row}: A genuine farmland-to-building conversion, where heatmap activation closely aligns with the annotated region. Such non-structural visual fluctuations make it challenging to distinguish phenology-driven false alarms from true land-use conversions.}
  \label{fig:phenology}
\end{figure*}%
}

\newcommand{\FigureFOne}{%
\begin{figure*}[t]
  \centering
  \includegraphics[width=0.8\textwidth]{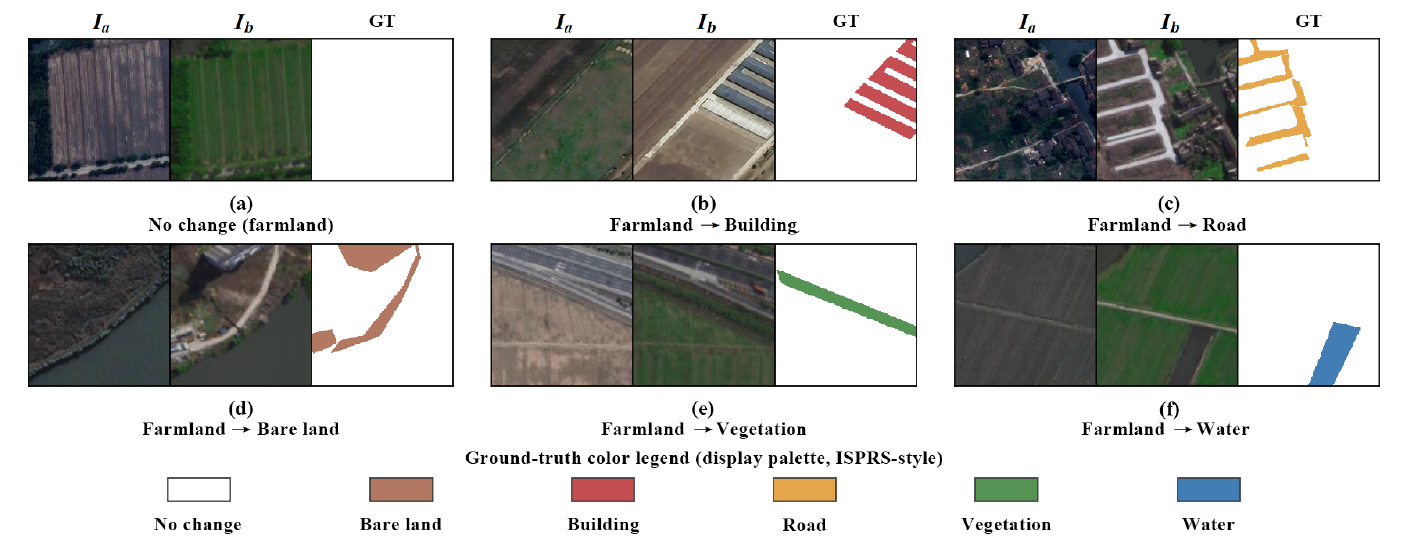} 
  \caption{Representative examples of the six semantic categories in HZNU-FCD. For each category, we show the bitemporal images ($I_{t_1}$ and $I_{t_2}$) and the corresponding pixel-level ``from-to'' annotation mask (GT). (a) No change (farmland). (b) Farmland $\rightarrow$ Building. (c) Farmland $\rightarrow$ Road. (d) Farmland $\rightarrow$ Bare land. (e) Farmland $\rightarrow$ Vegetation. (f) Farmland $\rightarrow$ Water.}
  \label{fig:dataset_samples}
\end{figure*}
}

\newcommand{\FigureFTwo}{
\begin{figure*}[h]
    \centering
    \includegraphics[width=1.0\textwidth]{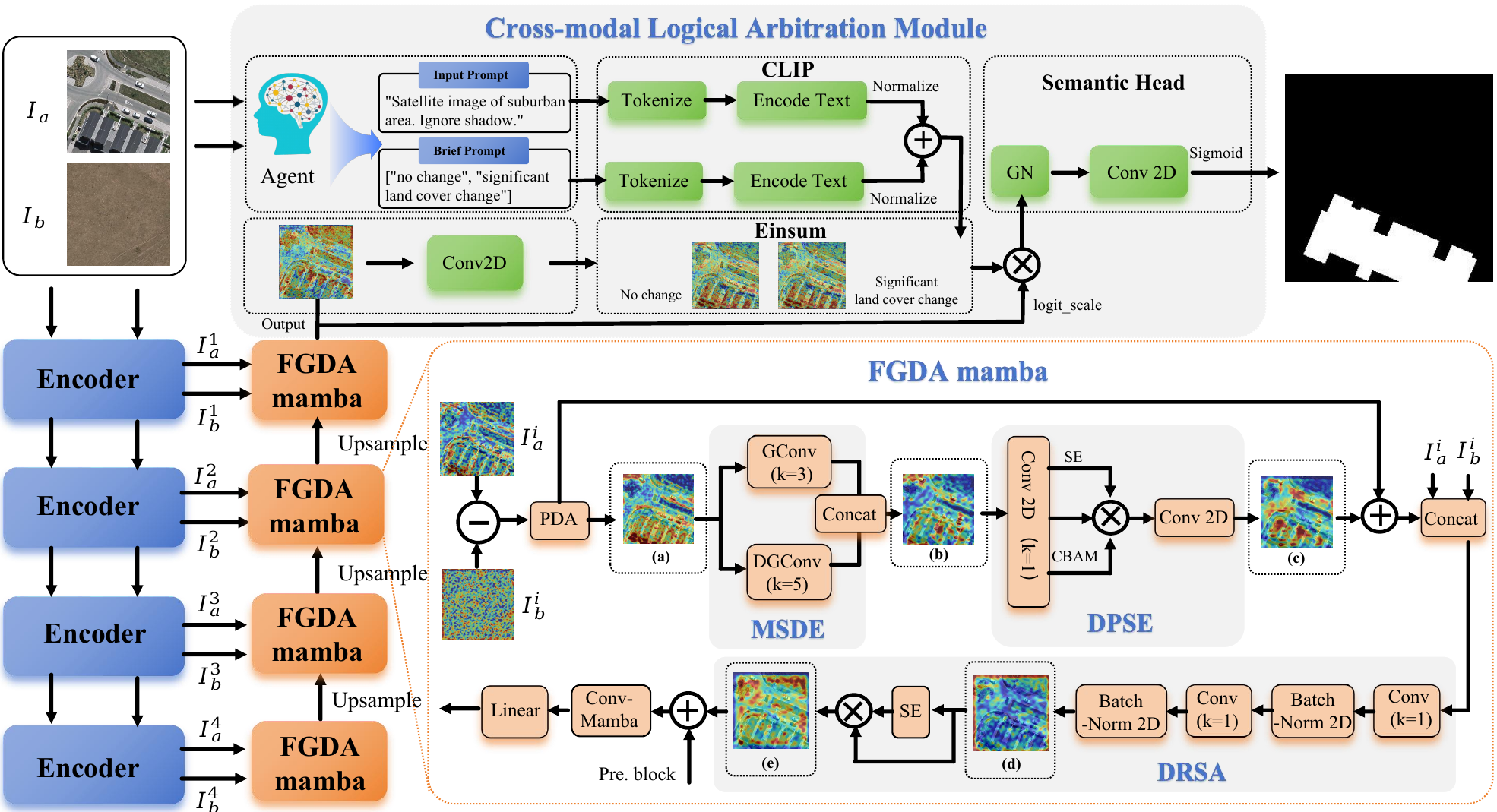}
    \vspace{-10pt}
    \caption{
Overall architecture of the proposed large-small collaborative framework for farmland SCD. The framework consists of a lightweight visual small model, FGDA-Mamba, and a large-model-driven CMLA module. FGDA-Mamba extracts fine-grained change-aware representations from bitemporal images through SRCM-Mamba encoding and FGDA-Mamba refinement. Inside the FGDA-Mamba block, PDA, MSDE, DPSE, DRSA, and Conv-Mamba progressively enhance reliable change cues, suppress pseudo-change responses, and preserve farmland boundary structures. The intermediate feature maps illustrate this refinement process: (a) initial difference features with visible but noisy structural responses; (b) multi-scale excavated features with more continuous change regions; (c) purified features after spatial-channel recalibration; (d) compact features after redundancy suppression; and (e) final Conv-Mamba-refined features with stronger focus on real changed regions. The CMLA module introduces CLIP-based textual priors from input and brief prompts, performs pixel-wise vision-language matching, and generates semantic gates to suppress non-semantic pseudo-changes and guide the final prediction.
}
\label{fig:overall}
    \vspace{-10pt}
\end{figure*}
}

\newcommand{\FigureFThree}{
\begin{figure}[h]
    \centering
    \includegraphics[width=0.5\textwidth]{Fig/F3.pdf}
    \vspace{-10pt}
    \caption{
Illustration of the LGSG module. Text prompts are encoded by the frozen CLIP text encoder to generate semantic prototypes, while the decoder output feature $Z$ is projected into the same embedding space. Pixel-wise vision-language similarities are computed to produce semantic responses, which are further scaled and used to guide the final output. LGSG introduces language priors to suppress pseudo-changes and enhance semantically consistent real change regions.
}
\label{fig:lgsg}
    \vspace{-10pt}
\end{figure}
}

\newcommand{\FigureArch}{
\begin{figure*}[h]
    \centering
    \includegraphics[width=1.0\textwidth]{Fig/F3.pdf}
    \vspace{-10pt}
    \caption{
Overall architecture of the proposed FGDA-Mamba framework. The network takes a pair of bitemporal farmland remote sensing images as input and follows an encoder--decoder structure. The SRCM-Mamba encoder extracts multi-scale bitemporal features, while the FGDA-Mamba Decoder progressively constructs fine-grained change-aware representations through PDA, MEDE, DPSE, DRSA, and Conv-Mamba reconstruction. The LGSG module further introduces CLIP-based textual priors for semantic consistency verification and produces the final semantic change map. The intermediate feature visualizations illustrate the progressive refinement process inside the FGDA-Mamba block. (a) The initial difference feature extracted by PDA, where building outlines and road structures are visible, but the responses are scattered and contain rich texture noise, making the contrast between real changes and background insufficient. (b) The feature map after MEDE, where multi-scale convolutions aggregate local details into more continuous regions, leading to smoother responses and clearer building boundaries. (c) The feature map after DPSE, where some change-related responses are strengthened and irrelevant regions are suppressed through spatial-channel recalibration. (d) The intermediate feature after DRSA, where the responses become more uniform and texture details are further standardized after dimensionality reduction and redundancy suppression. (e) The final feature after Conv-Mamba refinement, where warm responses are more concentrated on real changed buildings, while background regions maintain low activations, showing the strongest discriminative ability.
}
\label{fig:overall_model}
    \vspace{-10pt}
\end{figure*}
}

\newcommand{\FigureQualHZNU}{%
\begin{figure*}[!t]
\centering
\includegraphics[width=0.8\textwidth]{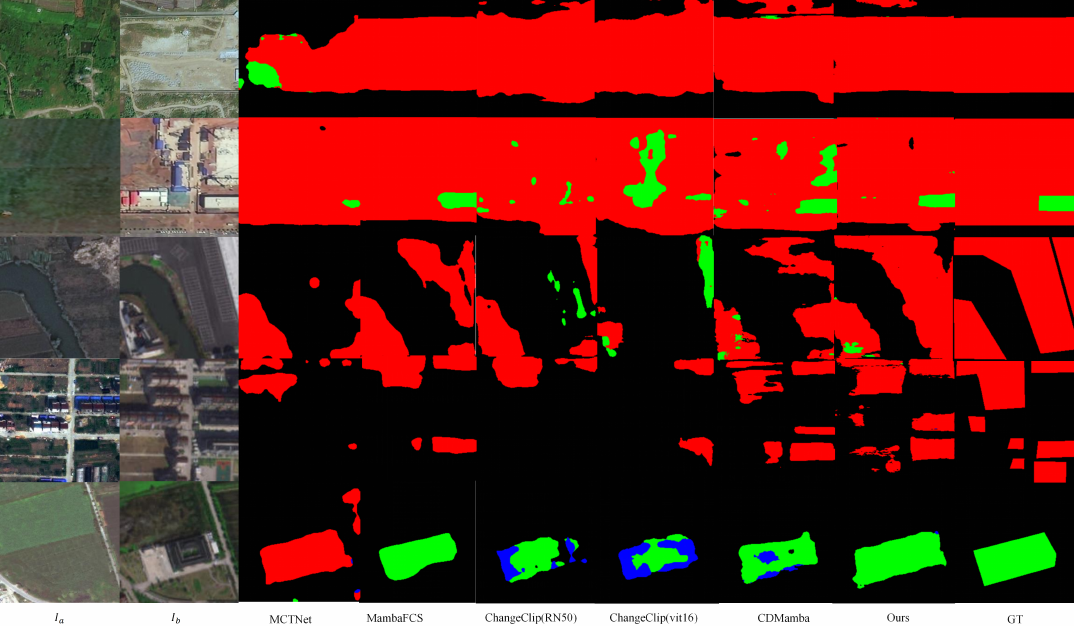}
\caption{Qualitative comparison on the HZNU-FCD test set. Color legend: red represents farmland$\rightarrow$bareland, green represents farmland$\rightarrow$building, blue represents farmland$\rightarrow$road, yellow represents farmland$\rightarrow$vegetation, and magenta represents farmland$\rightarrow$water.}
\label{fig:qual_hznu}
\end{figure*}%
}

\newcommand{\FigureQualLEVIR}{%
\begin{figure*}[!t]
\centering
\includegraphics[width=0.8\textwidth]{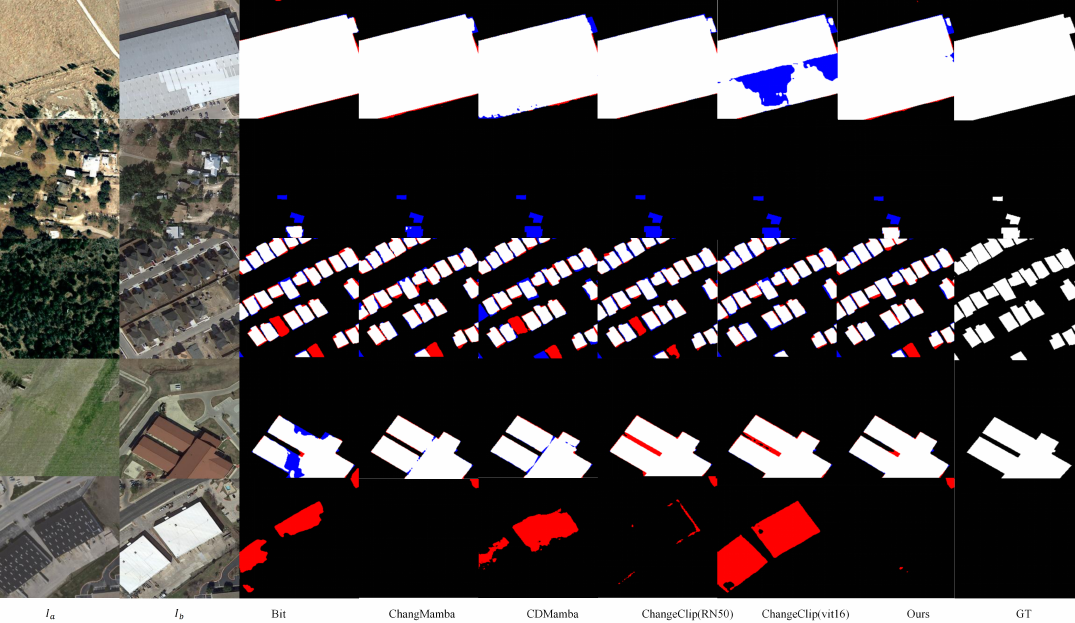}
\caption{Qualitative comparison on the LEVIR-CD test set. White denotes correctly detected change (TP), red denotes false positives (FP), blue denotes false negatives (FN), and black denotes background (TN).}
\label{fig:qual_levir}
\end{figure*}%
}

\newcommand{\FigureQualWHU}{%
\begin{figure*}[!t]
\centering
\includegraphics[width=0.8\textwidth]{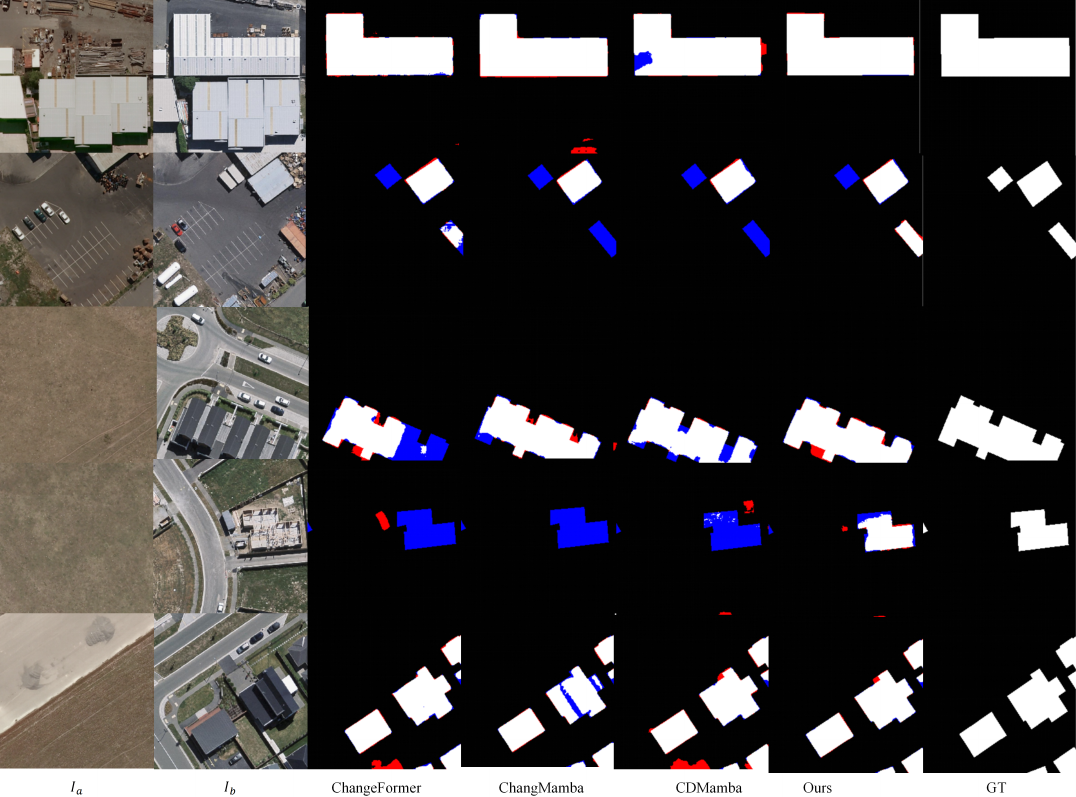}
\caption{Qualitative comparison on the WHU-CD test set. White denotes correctly detected change (TP), red denotes false positives (FP), blue denotes false negatives (FN), and black denotes background (TN).}
\label{fig:qual_whu}
\end{figure*}%
}

\newcommand{\FigureAblCMLABCDLEVIR}{%
\begin{figure}[!t]
\centering
\includegraphics[width=0.5\textwidth]{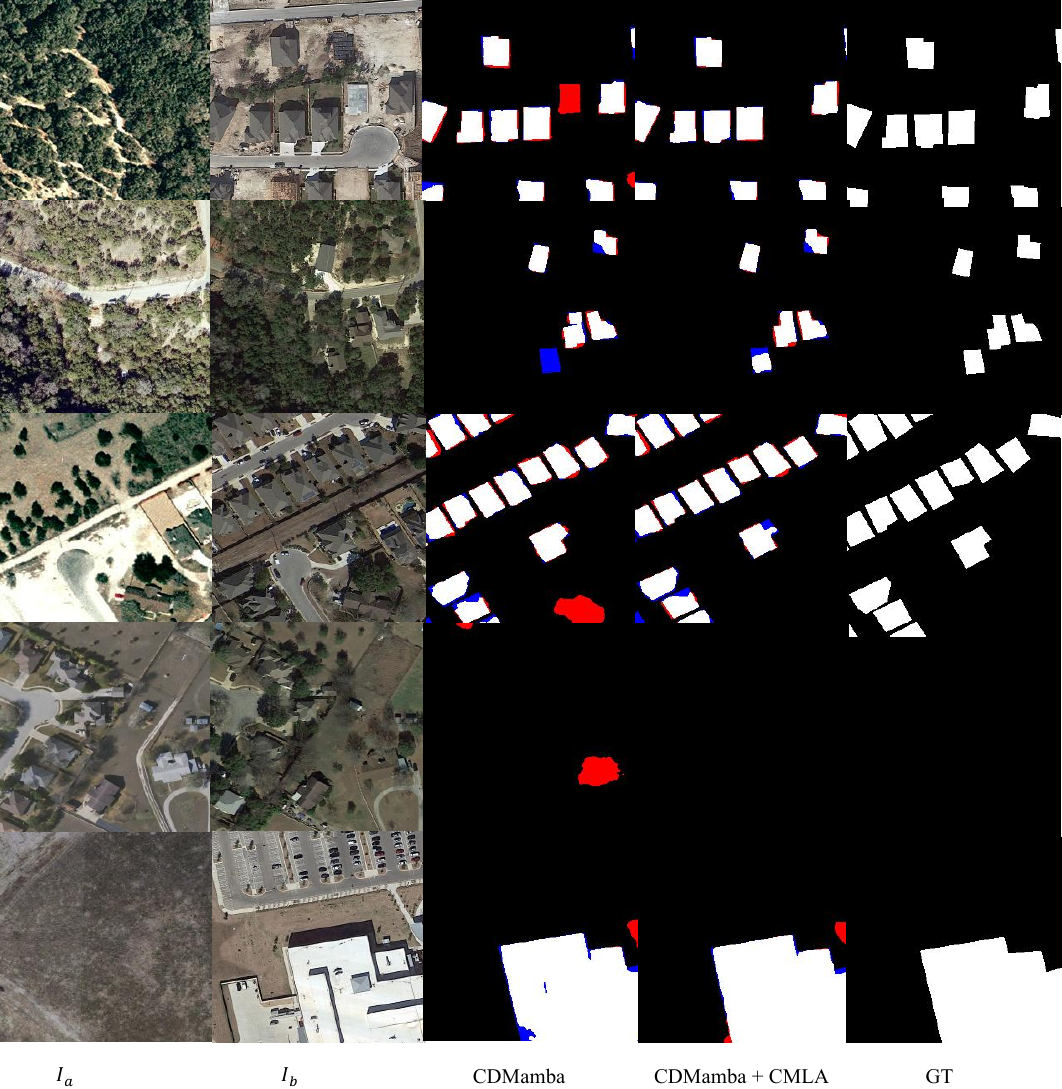}
\caption{Qualitative ablation of the CMLA module on LEVIR-CD. Red denotes false positives, and blue denotes false negatives.}
\label{fig:abl_CMLA_bcd_levir}
\end{figure}%
}

\newcommand{\FigureAblCMLABCDWHU}{%
\begin{figure}[!t]
\centering
\includegraphics[width=0.5\textwidth]{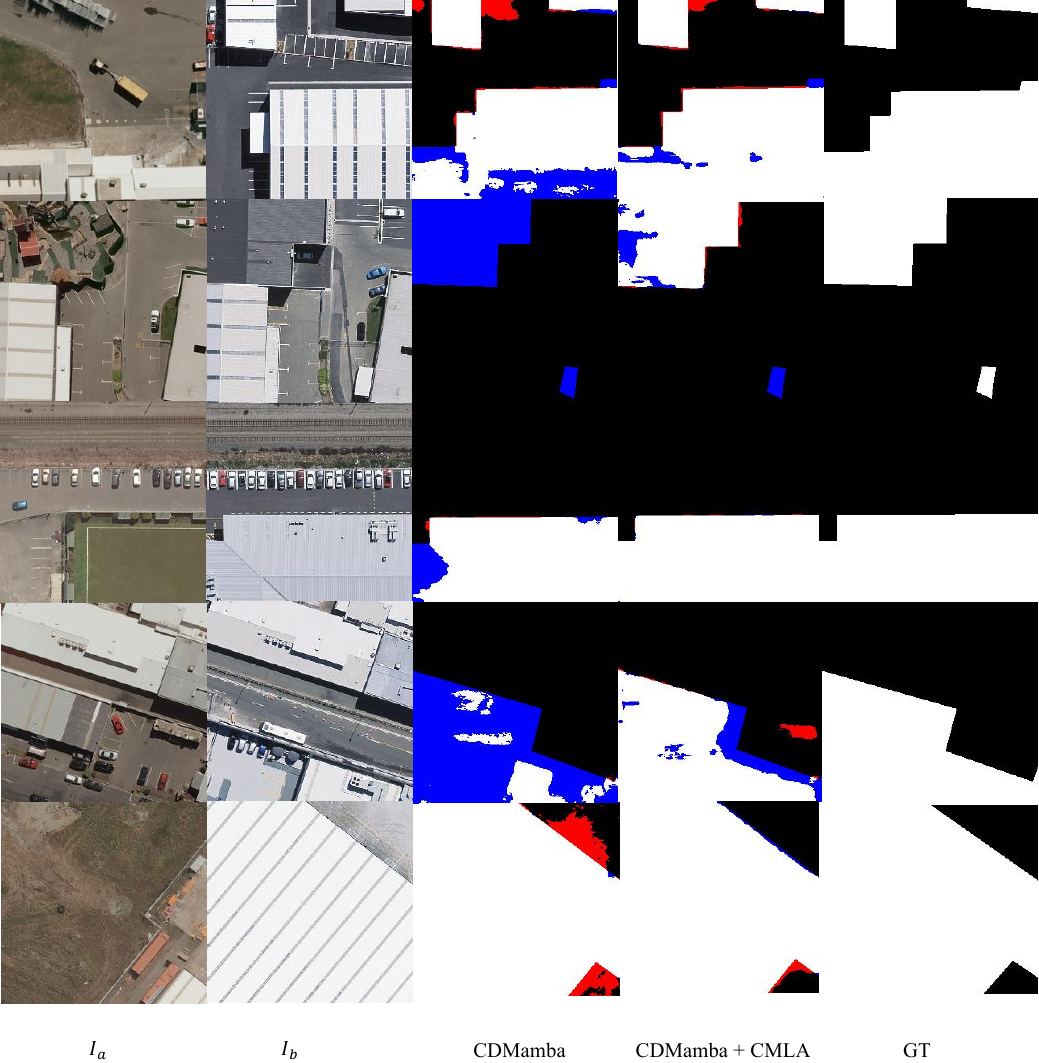}
\caption{Qualitative ablation of the CMLA module on WHU-CD. Red denotes false positives, and blue denotes false negatives.}
\label{fig:abl_CMLA_bcd_whu}
\end{figure}%
}

\newcommand{\FigureAblCMLASCD}{%
\begin{figure}[!t]
\centering
\includegraphics[width=0.5\textwidth]{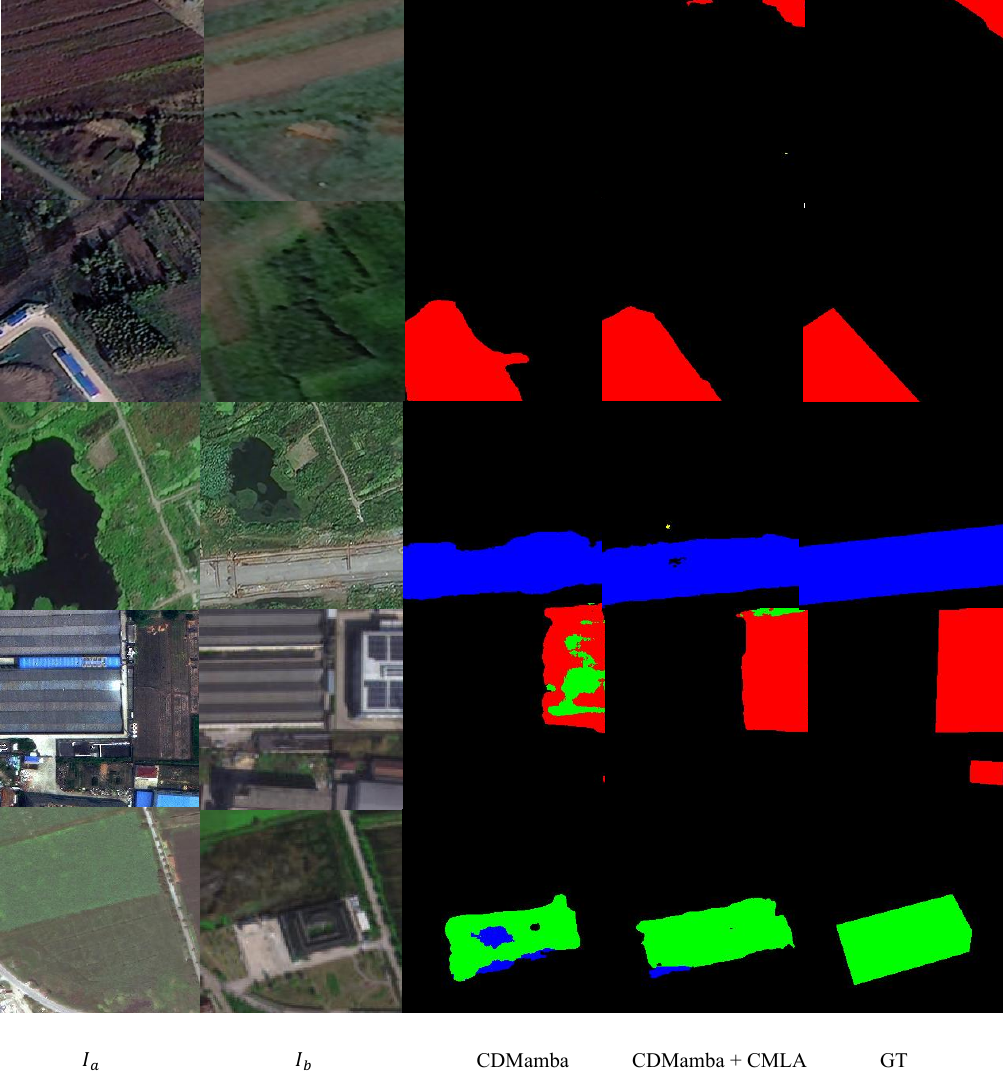}
\caption{Qualitative ablation of the CMLA module on the HZNU-FCD test set. Red represents farmland$\rightarrow$bareland, green represents farmland$\rightarrow$building, blue represents farmland$\rightarrow$road, yellow represents farmland$\rightarrow$vegetation, and magenta represents farmland$\rightarrow$water.}
\label{fig:abl_CMLA_scd}
\end{figure}%
}

\newcommand{\FigureFA}{%
\begin{figure}[t]
  \centering
  \includegraphics[width=\linewidth]{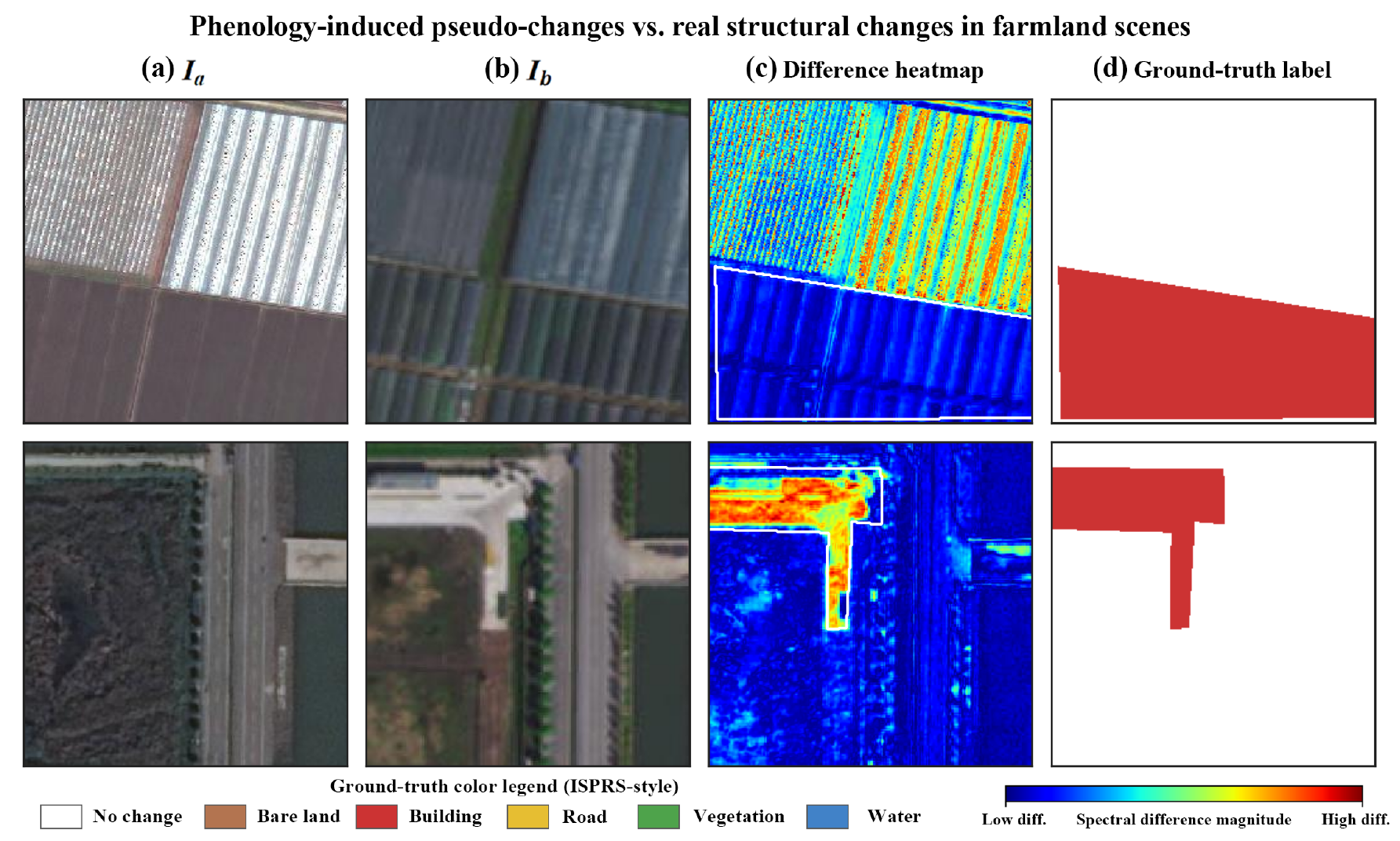}
  \caption{
Illustration of pseudo-changes and real structural changes in HZNU-FCD. 
(a)--(b) Bitemporal images $T_1$ and $T_2$; 
(c) pixel-wise spectral difference heatmap, where warm colors indicate larger spectral differences and the white contour marks the true change boundary; 
(d) ground-truth semantic label. 
The top row shows strong phenology-induced spectral shifts in unchanged farmland, while the bottom row shows a real farmland-to-building conversion. 
}
  \label{fig:phenology}
\end{figure}
}

%% file: table.tex
\newcommand{\TabHZNUFCDCompare}{%
\begin{table*}[!t]
\centering
\caption{Quantitative comparison on the HZNU-FCD test set. The best result in each metric is in \textbf{bold}, and the second best is \underline{underlined}.}
\label{tab:hznu_fcd_compare}
\renewcommand{\arraystretch}{1.15}
\setlength{\tabcolsep}{6pt}
\small
\begin{tabular*}{\textwidth}{@{\extracolsep{\fill}}lccccccc}
\toprule
\textbf{Method} & \textbf{Pre.(\%)} & \textbf{Rec.(\%)} & \textbf{F1(\%)} & \textbf{IoU(\%)} & \textbf{OA(\%)} & \textbf{$F_{\mathrm{scd}}$(\%)} & \textbf{SCD\_IoU$_{\mathrm{mean}}$(\%)} \\
\midrule
\multicolumn{8}{@{}l}{\textit{CNN-based}} \\
FC-EF~\cite{Dau18b}           & 75.63 & 72.23 & 73.89 & 58.59 & 83.12 & 73.89 & 73.76 \\
SNUNet~\cite{Fan21}     & 78.75 & 80.60 & 79.66 & 66.20 & 86.29 & 79.66 & 75.90 \\
\midrule
\multicolumn{8}{@{}l}{\textit{Transformer-based}} \\
ScratchFormer~\cite{Nom24}    & 81.55 & 78.65 & 80.08 & 66.77 & 85.72 & 74.95 & 73.38 \\
BiT-CD~\cite{Che21}           & 73.67 & 81.84 & 77.54 & 63.32 & 85.88 & 77.54 & 75.70 \\
CdSC~\cite{Wan24c}            & 94.41 & 93.27 & 93.84 & 88.39 & 95.53 & 92.80 & 90.81 \\
\midrule
\multicolumn{8}{@{}l}{\textit{Hybrid CNN--Transformer}} \\
SCanNet~\cite{SCanNet}        & 76.35 & \underline{97.23} & 85.54 & 74.73 & 88.01 & 83.49 & 78.07 \\
MCTNet~\cite{MCTNet}          & 95.57 & 88.75 & 92.03 & 85.24 & 94.75 & 92.03 & 90.98 \\
\midrule
\multicolumn{8}{@{}l}{\textit{Mamba-based}} \\
MambaFCS~\cite{MambaFCS}      & \underline{96.27} & 96.63 & 96.45 & 93.14 & 97.69 & 96.45 & 95.71 \\
ChangeMamba~\cite{Che24h}     & 95.56 & 96.44 & 96.00 & 92.31 & 97.44 & 96.00 & 95.42 \\
CDMamba~\cite{Zha24k}         & \textbf{97.85} & 97.16 & \underline{97.50} & \underline{96.10} & \underline{98.19} & \textbf{96.92} & \underline{96.16} \\
GSTM-SCD~\cite{Liu25g}        & 72.15 & 94.22 & 81.72 & 69.09 & 88.55 & 81.72 & 77.66 \\
\midrule
\multicolumn{8}{@{}l}{\textit{MultiModal-based}} \\
ChangeCLIP-RN50~\cite{Shi24a}  & 88.58 & 84.86 & 86.68 & 76.49 & 90.49 & 82.70 & 81.35 \\
ChangeCLIP-ViT~\cite{Shi24a} & 86.95 & 87.93 & 87.44 & 77.68 & 90.79 & 84.66 & 82.06 \\
\midrule
\textbf{Ours}    & \textbf{97.85} & \textbf{97.40} & \textbf{97.63} & \textbf{96.32} & \textbf{98.28} & \underline{96.87} & \textbf{96.35} \\
\bottomrule
\end{tabular*}
\end{table*}%
}

\newcommand{\TabLEVIRCompare}{%
\begin{table}[!t]
\centering
\caption{Quantitative comparison on the LEVIR-CD test set. The best result is in \textbf{bold}, and the second best is \underline{underlined}.}
\label{tab:levir_compare}
\renewcommand{\arraystretch}{1.15}
\setlength{\tabcolsep}{3pt}
\small
\begin{tabular}{lccccc}
\toprule
\textbf{Method} & \textbf{Pre.} & \textbf{Rec.} & \textbf{F1} & \textbf{IoU} & \textbf{OA} \\
\midrule
\multicolumn{6}{l}{\textit{CNN-based}} \\
FC-EF~\cite{Dau18b}            & 90.64 & 87.23 & 88.90 & 80.03 & 98.89 \\
FC-Siam-Diff~\cite{Dau18b}     & 90.81 & 88.59 & 89.69 & 81.31 & 98.96 \\
FC-Siam-Conc~\cite{Dau18b}     & 91.41 & 88.43 & 89.89 & 81.64 & 98.98 \\
IFNet~\cite{Zha20b}            & 89.62 & 86.65 & 88.11 & 78.75 & 98.81 \\
SNUNet~\cite{Fan21}           & 89.73 & 87.47 & 88.59 & 79.51 & 98.85 \\
\midrule
\multicolumn{6}{l}{\textit{Transformer-based}} \\
SwinUnet~\cite{Zha22f}         & 89.11 & 86.47 & 87.77 & 78.21 & 98.77 \\
BiT~\cite{Che21}              & 92.07 & 88.08 & 90.03 & 81.87 & 99.01 \\
ChangeFormer~\cite{Ban22}     & 90.68 & 87.04 & 88.83 & 79.90 & 98.88 \\
MSCANet~\cite{MSCANet}          & 90.02 & 88.71 & 89.36 & 80.77 & 98.92 \\
PaFormer~\cite{PaFormer}         & 91.34 & 88.07 & 89.68 & 81.29 & 98.96 \\
DARNet~\cite{DARNet}           & 92.19 & 88.99 & 90.56 & 82.76 & 99.05 \\
ACABFNet~\cite{ACABFNet}         & 90.11 & 88.27 & 89.18 & 80.48 & 98.91 \\
\midrule
\multicolumn{6}{l}{\textit{Mamba-based}} \\
RS-Mamba~\cite{Zha24l}         & 91.36 & 88.23 & 89.77 & 81.44 & 98.97 \\
ChangeMamba~\cite{Che24h}      & 91.59 & 88.78 & 90.16 & 82.09 & 99.01 \\
CDMamba~\cite{Zha24k}          & 91.43 & \underline{90.08} & 90.75 & 83.07 & \underline{99.06} \\
\midrule
\multicolumn{6}{l}{\textit{Multimodal-based}} \\
ChangeCLIP-ViT~\cite{Shi24a}   & \textbf{93.68} & 89.04 & \underline{91.30} & \underline{83.99} & \textbf{99.14} \\
\midrule
\textbf{Ours} & \underline{92.69} & \textbf{90.20} & \textbf{91.43} & \textbf{84.21} & \textbf{99.14} \\
\bottomrule
\end{tabular}
\end{table}%
}

\newcommand{\TabWHUCompare}{%
\begin{table}[!t]
\centering
\caption{Quantitative comparison on the WHU-CD test set. ``/'' indicates results that are not successfully reproduced. The best result is in \textbf{bold}, and the second best is \underline{underlined}.}
\label{tab:whu_compare}
\renewcommand{\arraystretch}{1.15}
\setlength{\tabcolsep}{3pt}
\small
\begin{tabular}{lccccc}
\toprule
\textbf{Method} & \textbf{Pre.} & \textbf{Rec.} & \textbf{F1} & \textbf{IoU} & \textbf{OA} \\
\midrule
\multicolumn{6}{l}{\textit{CNN-based}} \\
FC-EF~\cite{Dau18b}            & 92.10 & 90.64 & 91.36 & 84.10 & 99.32 \\
FC-Siam-Diff~\cite{Dau18b}     & 87.39 & 92.36 & 89.81 & 81.50 & 99.16 \\
FC-Siam-Conc~\cite{Dau18b}     & 86.57 & 91.11 & 88.78 & 79.83 & 99.08 \\
IFNet~\cite{Zha20b}            & 91.51 & 88.01 & 89.73 & 81.37 & 99.20 \\
SNUNet~\cite{Fan21}           & 84.70 & 89.73 & 87.14 & 77.22 & 98.95 \\
\midrule
\multicolumn{6}{l}{\textit{Transformer-based}} \\
SwinUnet~\cite{Zha22f}         & 92.44 & 87.56 & 89.93 & 81.71 & 99.22 \\
BiT~\cite{Che21}              & 91.84 & 91.95 & 91.90 & 85.01 & 99.35 \\
ChangeFormer~\cite{Ban22}     & 93.73 & 87.11 & 90.30 & 82.32 & 99.26 \\
MSCANet~\cite{MSCANet}          & 93.47 & 89.16 & 91.27 & 83.94 & 99.32 \\
PaFormer~\cite{PaFormer}         & 94.28 & 90.38 & 92.29 & 85.69 & 99.40 \\
DARNet~\cite{DARNet}           & 91.99 & 91.17 & 91.58 & 84.46 & 99.33 \\
ACABFNet~\cite{ACABFNet}         & 91.57 & 90.86 & 91.21 & 83.84 & 99.31 \\
\midrule
\multicolumn{6}{l}{\textit{Mamba-based}} \\
RS-Mamba~\cite{Zha24l}         & 95.50 & 90.24 & 92.79 & 86.55 & 99.44 \\
ChangeMamba~\cite{Che24h}      & 94.21 & 90.94 & 92.55 & 86.13 & 99.42 \\
CDMamba~\cite{Zha24k}          & \textbf{95.58} & 92.01 & \underline{93.76} & \underline{88.26} & \underline{99.51} \\
\midrule
\textbf{Ours} & \underline{94.97} & \textbf{92.75} & \textbf{93.85} & \textbf{88.41} & \textbf{99.52} \\
\bottomrule
\end{tabular}
\end{table}%
}

\newcommand{\TabAblFGDAComponents}{%
\begin{table}[!t]
\centering
\caption{Component ablation of the FGDA-Mamba Decoder on the HZNU-FCD test set. \checkmark~denotes the submodule is enabled. The best result in each column is in \textbf{bold}. }
\label{tab:abl_fgda_components}
\renewcommand{\arraystretch}{1.15}
\setlength{\tabcolsep}{5pt}
\small
\begin{tabular}{ccc|ccc}
\toprule
\textbf{MEDE} & \textbf{DPSE} & \textbf{DRSA} & \textbf{F1(\%)} & \textbf{IoU(\%)} & \textbf{$F_{\mathrm{scd}}$(\%)} \\
\midrule
            &             &             & 96.663 & 93.541 & 95.635 \\
            & \checkmark  & \checkmark  & 97.359 & 94.853 & 96.259 \\
\checkmark  &             & \checkmark  & 96.845 & 93.883 & 95.900 \\
\checkmark  & \checkmark  &             & 97.528 & 95.175 & 96.721 \\
\midrule
\checkmark  & \checkmark  & \checkmark  & \textbf{97.630} & \textbf{96.320} & \textbf{96.870} \\
            &             &             & \scriptsize{(+0.967)} & \scriptsize{(+2.779)} & \scriptsize{(+1.235)} \\
\bottomrule
\end{tabular}
\end{table}%
}

\newcommand{\TabAblCMLABCD}{%
\begin{table}[!t]
\centering
\caption{Ablation study of the CMLA module on the binary change detection benchmarks LEVIR-CD and WHU-CD. }
\label{tab:abl_CMLA_bcd}
\renewcommand{\arraystretch}{1.15}
\setlength{\tabcolsep}{3.5pt}
\small
\begin{tabular}{llccc}
\toprule
\textbf{Dataset} & \textbf{Setting} & \textbf{F1} & \textbf{IoU} & \textbf{OA} \\
\midrule
\multirow{2}{*}{LEVIR-CD}
 & CDMamba       & 90.99 & 83.47 & 99.25 \\
 & CDMamba+CMLA  & \textbf{91.74} & \textbf{84.75} & \textbf{99.32} \\
\midrule
\multirow{2}{*}{WHU-CD}
 & CDMamba       & 92.88 & 86.71 & 99.32 \\
 & CDMamba+CMLA  & \textbf{94.38} & \textbf{89.35} & \textbf{99.45} \\
\bottomrule
\end{tabular}
\end{table}%
}

\newcommand{\TabAblCMLASCD}{%
\begin{table}[!t]
\centering
\caption{Ablation study of the CMLA module on the farmland-oriented semantic change detection dataset HZNU-FCD. }
\label{tab:abl_CMLA_scd}
\renewcommand{\arraystretch}{1.15}
\setlength{\tabcolsep}{3.0pt}
\small
\begin{tabular}{lcccc}
\toprule
\textbf{Setting} & \textbf{F1} & \textbf{IoU} & \textbf{OA} & \textbf{$F_{\mathrm{scd}}$} \\
\midrule
CDMamba      & 97.50 & 96.10 & 98.19 & \textbf{96.92} \\
CDMamba+CMLA & \textbf{97.53} & \textbf{96.19} & \textbf{98.28} & 96.62 \\
\bottomrule
\end{tabular}
\end{table}%
}

\newcommand{\TabAblTauLambda}{%
\begin{table}[!t]
\centering
\caption{Hyperparameter analysis of the hard-mask auxiliary loss on the LEVIR-CD test set. $\tau$ denotes the hard-pixel threshold and $\lambda$ denotes the auxiliary weight.}
\label{tab:abl_tau_lambda}
\renewcommand{\arraystretch}{1.15}
\setlength{\tabcolsep}{4pt}
\small
\begin{tabular}{lccc}
\toprule
\textbf{Setting} & \textbf{F1(\%)} & \textbf{IoU(\%)} & \textbf{OA(\%)} \\
\midrule
$\tau{=}0.80, \lambda{=}0.30$ & 91.40 & 84.15 & 99.14 \\
$\tau{=}0.80, \lambda{=}0.40$ & \textbf{91.43} & \textbf{84.21} & \textbf{99.14} \\
$\tau{=}0.80, \lambda{=}0.50$ & 91.27 & 83.91 & 99.12 \\
$\tau{=}0.80, \lambda{=}0.60$ & 91.26 & 83.93 & 99.12 \\
$\tau{=}0.85, \lambda{=}0.40$ & 91.35 & 84.08 & 99.12 \\
$\tau{=}0.75, \lambda{=}0.40$ & 91.19 & 83.80 & 99.12 \\
\bottomrule
\end{tabular}
\end{table}%
}

\newcommand{\TabDatasetComparison}{%
\begin{table}[t]
\centering
\small
\setlength{\tabcolsep}{4pt}
\caption{Comparison of farmland-related change detection datasets. ``SCD'' = semantic CD; ``BCD'' = binary CD.}
\label{tab:dataset_comparison}
\renewcommand{\arraystretch}{1.15}
\begin{tabular}{lcccc}
\toprule
Dataset & Task & \#Pairs & Classes & Source \\
\midrule
CLCD~\cite{Liu22clcd}                  & BCD & 600         & --           & GF-2 \\
JL1-CD~\cite{JL1Cup2024}  & BCD & 5,000       & --           & Jilin-1 \\
JL1 Cup SCD~\cite{JL1Cup2024}   & SCD & $\sim$8,000 & 8 bidir.     & Jilin-1 \\
Hi-CNA~\cite{Sun24b}  & SCD & 6,797       & 4 coarse     & GF-2 \\
\textbf{HZNU-FCD (Ours)} & \textbf{SCD} & \textbf{4,588} & \textbf{5 fine$^{\dagger}$} & \textbf{UAV+Jilin-1} \\
\bottomrule
\end{tabular}\\[3pt]
\footnotesize
$^{\dagger}$The five classes in HZNU-FCD are \textit{farmland-to-non-farmland} conversion types (building, road, bareland, vegetation, water), specifically designed for cultivated land protection and law enforcement. Unlike general multi-directional SCD datasets, all annotations take farmland as the exclusive $T_{1}$ reference.
\end{table}
}

%% file: references.bib
@inproceedings{Dau18b,
  author={Daudt, R. C. and Le Saux, B. and Boulch, A.},
  title={Fully Convolutional Siamese Networks for Change Detection},
  booktitle={2018 25th IEEE International Conference on Image Processing (ICIP)},
  year={2018},
  pages={4063--4067},
  doi={10.1109/ICIP.2018.8451652}
}

@article{Zha17,
  author={Zhan, Y. and Fu, K. and Yan, M. and Sun, X. and Wang, H. and Qiu, X.},
  title={Change Detection Based on Deep Siamese Convolutional Network for Optical Aerial Images},
  journal={IEEE Geoscience and Remote Sensing Letters},
  year={2017},
  volume={14},
  number={10},
  pages={1845--1849},
  doi={10.1109/LGRS.2017.2738149}
}

@article{Zha20b,
  author={Zhang, C. and Yue, P. and Tapete, D. and Jiang, L. and Shangguan, B. and Huang, L. and Liu, G.},
  title={A Deeply Supervised Image Fusion Network for Change Detection in High Resolution Bi-Temporal Remote Sensing Images},
  journal={ISPRS Journal of Photogrammetry and Remote Sensing},
  year={2020},
  volume={166},
  pages={183--200},
  doi={10.1016/j.isprsjprs.2020.06.003}
}

@article{Fan21,
  author={Fang, S. and Li, K. and Shao, J. and Li, Z.},
  title={SNUNet-CD: A Densely Connected Siamese Network for Change Detection of VHR Images},
  journal={IEEE Geoscience and Remote Sensing Letters},
  year={2021},
  volume={19},
  pages={1--5},
  doi={10.1109/LGRS.2021.3056416}
}

@article{Shi21,
  author={Shi, Q. and Liu, M. and Li, S. and Liu, X. and Wang, F. and Zhang, L.},
  title={A Deeply Supervised Attention Metric-Based Network and an Open Aerial Image Dataset for Remote Sensing Change Detection},
  journal={IEEE Transactions on Geoscience and Remote Sensing},
  year={2022},
  volume={60},
  pages={1--16},
  doi={10.1109/TGRS.2021.3085870}
}

@article{Che21,
  author={Chen, H. and Qi, Z. and Shi, Z.},
  title={Remote Sensing Image Change Detection With Transformers},
  journal={IEEE Transactions on Geoscience and Remote Sensing},
  year={2022},
  volume={60},
  pages={1--14},
  doi={10.1109/TGRS.2021.3095166}
}

@inproceedings{Ban22,
  author={Bandara, W. G. C. and Patel, V. M.},
  title={A Transformer-Based Siamese Network for Change Detection},
  booktitle={2022 IEEE International Geoscience and Remote Sensing Symposium (IGARSS)},
  year={2022},
  pages={207--210},
  doi={10.1109/IGARSS46834.2022.9883686}
}

@article{Zha22f,
  author={Zhang, C. and Wang, L. and Cheng, S. and Li, Y.},
  title={SwinSUNet: Pure Transformer Network for Remote Sensing Image Change Detection},
  journal={IEEE Transactions on Geoscience and Remote Sensing},
  year={2022},
  volume={60},
  pages={1--13},
  doi={10.1109/TGRS.2022.3160007}
}

@article{Zha23j,
  author={Zhang, K. and Luppino, L. T. and Zhu, X. X. and Bruzzone, L.},
  title={Relation Changes Matter: Cross-Temporal Difference Transformer for Change Detection in Remote Sensing Images},
  journal={IEEE Transactions on Geoscience and Remote Sensing},
  year={2023},
  volume={61},
  pages={1--15},
  doi={10.1109/TGRS.2023.3281711}
}

@article{Lei24,
  author={Lei, T. and Xu, Y. and Ning, H. and Lv, Z. and Min, C. and Jin, Y. and Nandi, A. K.},
  title={Lightweight Structure-Aware Transformer Network for Remote Sensing Image Change Detection},
  journal={IEEE Geoscience and Remote Sensing Letters},
  year={2024},
  volume={21},
  pages={1--5},
  number={6000305},
  doi={10.1109/LGRS.2023.3323534}
}

@article{Che24h,
  author={Chen, H. and Song, J. and Han, C. and Xia, J. and Yokoya, N.},
  title={ChangeMamba: Remote Sensing Change Detection With Spatiotemporal State Space Model},
  journal={IEEE Transactions on Geoscience and Remote Sensing},
  year={2024},
  volume={62},
  pages={1--20},
  doi={10.1109/TGRS.2024.3417253}
}

@article{Zha24l,
  author={Zhao, S. and Chen, H. and Zhang, X. and Xiao, P. and Bai, L. and Ouyang, W.},
  title={RS-Mamba for Large Remote Sensing Image Dense Prediction},
  journal={IEEE Transactions on Geoscience and Remote Sensing},
  year={2024},
  volume={62},
  pages={1--14},
  doi={10.1109/TGRS.2024.3425540}
}

@article{Zha24k,
  author={Zhang, H. and Chen, K. and Liu, C. and Chen, H. and Zou, Z. and Shi, Z.},
  title={CDMamba: Incorporating Local Clues Into Mamba for Remote Sensing Image Binary Change Detection},
  journal={IEEE Transactions on Geoscience and Remote Sensing},
  year={2025},
  volume={63},
  pages={1--16},
  doi={10.1109/TGRS.2025.3545012}
}

@article{Don24c,
  author={Dong, Z. and Yuan, G. and Hua, Z. and Li, J.},
  title={ConMamba: CNN and SSM High-Performance Hybrid Network for Remote Sensing Change Detection},
  journal={IEEE Transactions on Geoscience and Remote Sensing},
  year={2024},
  volume={62},
  pages={1--15},
  doi={10.1109/TGRS.2024.3495216}
}

@article{Gu23,
  author={Gu, A. and Dao, T.},
  title={Mamba: Linear-Time Sequence Modeling with Selective State Spaces},
  year={2023},
  eprint={2312.00752},
  archivePrefix={arXiv},
  primaryClass={cs.LG}
}

@article{Sun24b,
  author={Sun, Z. and Zhong, Y. and Wang, X. and Zhang, L.},
  title={Identifying Cropland Non-Agriculturalization With High Representational Consistency From Bi-Temporal High-Resolution Remote Sensing Images: From Benchmark Datasets to Real-World Application},
  journal={ISPRS Journal of Photogrammetry and Remote Sensing},
  year={2024},
  volume={212},
  pages={454--474},
  doi={10.1016/j.isprsjprs.2024.05.011}
}

@article{chen2020levir,
  author={Chen, H. and Shi, Z.},
  title={A Spatial-Temporal Attention-Based Method and a New Dataset for Remote Sensing Image Change Detection},
  journal={Remote Sensing},
  year={2020},
  volume={12},
  number={10},
  pages={1662},
  doi={10.3390/rs12101662}
}

@inproceedings{ji2018whu,
  author={Ji, S. and Wei, S. and Lu, M.},
  title={Fully Convolutional Networks for Multisource Building Extraction From an Open Aerial and Satellite Imagery Dataset},
  booktitle={IEEE Transactions on Geoscience and Remote Sensing},
  year={2018},
  volume={57},
  number={1},
  pages={574--586},
  doi={10.1109/TGRS.2018.2858817}
}

@inproceedings{Rad21,
  author={Radford, A. and Kim, J. W. and Hallacy, C. and Ramesh, A. and Goh, G. and Agarwal, S. and Sastry, G. and Askell, A. and Mishkin, P. and Clark, J. and Krueger, G. and Sutskever, I.},
  title={Learning Transferable Visual Models From Natural Language Supervision},
  booktitle={Proceedings of the International Conference on Machine Learning},
  year={2021},
  pages={8748--8763}
}

@article{Shi24a,
  author={Shi, W. and Zhang, M. and Zhang, R. and Chen, S. and Zhan, Z.},
  title={ChangeCLIP: Remote Sensing Change Detection with Sample-Efficient Vision-Language Semantic Alignment},
  journal={ISPRS Journal of Photogrammetry and Remote Sensing},
  year={2024},
  volume={208},
  pages={1--14},
  doi={10.1016/j.isprsjprs.2023.11.013}
}

@article{SemanticCD,
  author={Liu, Y. and Peng, D. and Zhang, X. and Guo, Q. and Zhong, Y. and Zhang, L.},
  title={Semantic Change Detection via Bidirectional Vision-Language Feature Alignment},
  journal={IEEE Transactions on Geoscience and Remote Sensing},
  year={2024},
  volume={62},
  pages={1--14},
  doi={10.1109/TGRS.2024.3381234}
}

@article{Yan22a,
  author={Yang, K. and Xia, G.-S. and Liu, Z. and Du, B. and Yang, W. and Pelillo, M. and Zhang, L.},
  title={Asymmetric Siamese Networks for Semantic Change Detection in Aerial Images},
  journal={IEEE Transactions on Geoscience and Remote Sensing},
  year={2022},
  volume={60},
  pages={1--18},
  doi={10.1109/TGRS.2021.3083818}
}

@article{Dau19,
  author={Daudt, R. C. and Le Saux, B. and Boulch, A. and Gousseau, Y.},
  title={Multitask Learning for Large-Scale Semantic Change Detection},
  journal={Computer Vision and Image Understanding},
  year={2019},
  volume={187},
  pages={102783},
  doi={10.1016/j.cviu.2019.07.003}
}

@article{Din22,
  author={Ding, L. and Guo, H. and Liu, S. and Mou, L. and Zhang, J. and Bruzzone, L.},
  title={{Bi-Temporal} Semantic Reasoning for the Semantic Change Detection in {HR} Remote Sensing Images},
  journal={IEEE Transactions on Geoscience and Remote Sensing},
  year={2022},
  volume={60},
  pages={1--14},
  doi={10.1109/TGRS.2022.3154368}
}

@article{MCTNet,
  author={Song, T. and Zhang, X. and Li, J. and Gao, L. and Li, B. and Peng, M.},
  title={{MCTNet}: Multi-Context Transformer Network for Semantic Change Detection of Remote Sensing Images},
  journal={IEEE Transactions on Geoscience and Remote Sensing},
  year={2023},
  volume={61},
  pages={1--15},
  doi={10.1109/TGRS.2023.3290261}
}

@article{SCanNet,
  author={Du, Y. and Xu, J. and Zhu, X. and Qiu, X. and Wei, Z.},
  title={{SCanNet}: Joint Spatiotemporal Convolutional-and-Attention Network for Semantic Change Detection},
  journal={IEEE Transactions on Geoscience and Remote Sensing},
  year={2023},
  volume={61},
  pages={1--15},
  doi={10.1109/TGRS.2023.3300199}
}

@article{HGINet,
  author={Zheng, Z. and Wan, Q. and Zhang, Y. and Zhong, Y. and Zhang, L.},
  title={{HGINet}: Hierarchical Graph Interaction Transformer with Edge-Indicated Attention for Semantic Change Detection},
  journal={IEEE Transactions on Geoscience and Remote Sensing},
  year={2024},
  volume={62},
  pages={1--16},
  doi={10.1109/TGRS.2024.3349908}
}

@article{CdSC,
  author={Ding, L. and Tang, H. and Bruzzone, L.},
  title={LandSCD: Change Detection Based on Change of Land Surface Characteristics Under Semantic Constraints},
  journal={IEEE Transactions on Geoscience and Remote Sensing},
  year={2022},
  volume={60},
  pages={1--14},
  doi={10.1109/TGRS.2022.3166465}
}

@article{MambaFCS,
  author={Zhao, Z. and Zhang, H. and He, Y. and Zhou, Y. and Shi, Z.},
  title={{MambaFCS}: Selective State Space Model with Frequency-Domain Cues for Remote Sensing Image Change Detection},
  journal={IEEE Transactions on Geoscience and Remote Sensing},
  year={2024},
  volume={62},
  pages={1--15},
  doi={10.1109/TGRS.2024.3484602}
}

@article{MSCANet,
  author={Chen, J. and Yuan, Z. and Peng, J. and Chen, L. and Han, H. and Chu, J. and Fan, X. and Li, H.},
  title={{IFNet}: Deep Fusion of Multi-Scale and Multi-Spectral Information for Change Detection in Optical and {SAR} Images},
  journal={IEEE Transactions on Geoscience and Remote Sensing},
  year={2022},
  volume={60},
  pages={1--13},
  doi={10.1109/TGRS.2021.3127540}
}

@article{PaFormer,
  author={Liu, X. and Li, Z. and Zhao, W. and Shi, J. and Zhou, J.},
  title={{PaFormer}: Parallel Attentional Transformer for Remote Sensing Change Detection},
  journal={IEEE Geoscience and Remote Sensing Letters},
  year={2023},
  volume={20},
  pages={1--5},
  doi={10.1109/LGRS.2023.3288991}
}

@article{DARNet,
  author={Mao, Z. and Zhong, Y. and Hu, X. and Cao, L. and Gao, J. and Zhang, L.},
  title={{DARNet}: Semantic Supervised Dense Attention Retargeting Network for Change Detection in Remote Sensing Images},
  journal={IEEE Transactions on Geoscience and Remote Sensing},
  year={2023},
  volume={61},
  pages={1--15},
  doi={10.1109/TGRS.2023.3291483}
}

@article{ACABFNet,
  author={Li, J. and Su, Z. and Geng, J. and Yin, Y.},
  title={{ACABFNet}: Attentional Class-Aware Background Features for Remote Sensing Image Change Detection},
  journal={IEEE Geoscience and Remote Sensing Letters},
  year={2022},
  volume={19},
  pages={1--5},
  doi={10.1109/LGRS.2022.3182209}
}

@article{Lun06,
  author={Lunetta, R. S. and Knight, J. F. and Ediriwickrema, J. and Lyon, J. G. and Worthy, L. D.},
  title={Land-Cover Change Detection Using Multi-Temporal {MODIS NDVI} Data},
  journal={Remote Sensing of Environment},
  year={2006},
  volume={105},
  number={2},
  pages={142--154},
  doi={10.1016/j.rse.2006.06.018}
}

@article{Nom24,
  author={Noman, M. and Fiaz, M. and Cholakkal, H. and Narayan, S. and Muhammad Anwer, R. and Khan, S. and Shahbaz Khan, F.},
  title={Remote Sensing Change Detection With Transformers Trained From Scratch},
  journal={IEEE Transactions on Geoscience and Remote Sensing},
  year={2024},
  volume={62},
  pages={1--14},
  doi={10.1109/TGRS.2024.3383800}
}

@article{Wan24c,
  author={Wang, Q. and Jing, W. and Chi, K. and Yuan, Y.},
  title={Cross-Difference Semantic Consistency Network for Semantic Change Detection},
  journal={IEEE Transactions on Geoscience and Remote Sensing},
  year={2024},
  volume={62},
  pages={1--12},
  doi={10.1109/TGRS.2024.3386334}
}

@article{Liu25g,
  author={Liu, X. and Dai, C. and Ding, L. and Zhang, Z. and Li, Y. and Zuo, X. and Li, M. and Wang, H. and Miao, Y.},
  title={{GSTM-SCD}: Graph-enhanced spatio-temporal state space model for semantic change detection in multi-temporal remote sensing images},
  journal={ISPRS Journal of Photogrammetry and Remote Sensing},
  year={2025},
  volume={230},
  pages={73--91},
  doi={10.1016/j.isprsjprs.2025.09.003}
}

@article{Liu22clcd,
  author  = {Liu, Mengxi and Chai, Zhuoqun and Deng, Haojun and Liu, Rong},
  journal = {IEEE Journal of Selected Topics in Applied Earth Observations and Remote Sensing},
  title   = {A {CNN}-Transformer Network With Multiscale Context Aggregation for Fine-Grained Cropland Change Detection},
  year    = {2022},
  volume  = {15},
  pages   = {4297--4306},
  doi     = {10.1109/JSTARS.2022.3177235}
}

@misc{JL1Cup2024,
  title        = {Jilin-1 Cup 2024 Remote Sensing Image Intelligent Processing Competition, Track 2: Farmland Semantic Change Detection},
  author       = {{Chang Guang Satellite Technology Co., Ltd.}},
  year         = {2024},
  url          = {https://www.jl1mall.com/contest/match/info?id=1645664411716952066},
  note         = {Accessed: 2026-05-09}
}

@inproceedings{chen2025multi,
  title={Multi-modal medical diagnosis via large-small model collaboration},
  author={Chen, Wanyi and Zhao, Zihua and Yao, Jiangchao and Zhang, Ya and Bu, Jiajun and Wang, Haishuai},
  booktitle={Proceedings of the Computer Vision and Pattern Recognition Conference},
  pages={30763--30773},
  year={2025}
}

@inproceedings{liu2024cotuning,
  title={Cotuning: A large-small model collaborating distillation framework for better model generalization},
  author={Liu, Zimo and Liu, Kangjun and Guo, Mingyue and Zhang, Shiliang and Wang, Yaowei},
  booktitle={Proceedings of the 32nd ACM International Conference on Multimedia},
  pages={10487--10496},
  year={2024}
}

@inproceedings{lu2024collaborative,
  title={Collaborative training of tiny-large vision language models},
  author={Lu, Shichen and Guo, Longteng and Wang, Wenxuan and Zhao, Zijia and Yue, Tongtian and Liu, Jing and Liu, Si},
  booktitle={Proceedings of the 32nd ACM International Conference on Multimedia},
  pages={4928--4937},
  year={2024}
}

@inproceedings{wang2026collaborative,
  title={Collaborative Enhancement of Large and Small Models for Question Answering via Dual Knowledge Transfer},
  author={Wang, Shaofei and Liu, Yunan and Tang, Xiaolan and Chen, Wenlong},
  booktitle={Proceedings of the AAAI Conference on Artificial Intelligence},
  volume={40},
  number={40},
  pages={33630--33638},
  year={2026}
}

@article{wang2022multimodal,
  title={Multimodal adaptive distillation for leveraging unimodal encoders for vision-language tasks},
  author={Wang, Zhecan and Codella, Noel and Chen, Yen-Chun and Zhou, Luowei and Dai, Xiyang and Xiao, Bin and Yang, Jianwei and You, Haoxuan and Chang, Kai-Wei and Chang, Shih-fu and others},
  journal={arXiv preprint arXiv:2204.10496},
  year={2022}
}

@inproceedings{chen2024data,
  title={Data shunt: Collaboration of small and large models for lower costs and better performance},
  author={Chen, Dong and Zhuang, Yueting and Zhang, Shuo and Liu, Jinfeng and Dong, Su and Tang, Siliang},
  booktitle={Proceedings of the AAAI Conference on Artificial Intelligence},
  volume={38},
  number={10},
  pages={11249--11257},
  year={2024}
}

@article{wijenayake2026mamba,
  title={Mamba-FCS: Joint Spatio-Frequency Feature Fusion, Change-Guided Attention, and Sek Inspired Loss for Enhanced Semantic Change Detection in Remote Sensing},
  author={Wijenayake, Buddhi and Ratnayake, Athulya and Sumanasekara, Praveen and Godaliyadda, Roshan and Ekanayake, Parakrama and Herath, Vijitha and Wasalathilaka, Nichula},
  journal={IEEE Journal of Selected Topics in Applied Earth Observations and Remote Sensing},
  year={2026},
  publisher={IEEE}
}

@article{qiu2024novel,
  title={A novel change detection method based on visual language from high-resolution remote sensing images},
  author={Qiu, Junlong and Liu, Wei and Zhang, Hui and Li, Erzhu and Zhang, Lianpeng and Li, Xing},
  journal={IEEE Journal of Selected Topics in Applied Earth Observations and Remote Sensing},
  volume={18},
  pages={4554--4567},
  year={2024},
  publisher={IEEE}
}

@article{wu2025remote,
  title={A Remote Sensing Image Change Detection Network With Feature Constraints From a Visual Foundation Model},
  author={Wu, Zhaoming and Zan, Luyang and Chen, Zhengchao and Cai, Mingyong and Li, Yixiang and Wang, Zeqing and Xie, Jun and Shi, Xuewei},
  journal={IEEE Journal of Selected Topics in Applied Earth Observations and Remote Sensing},
  volume={18},
  pages={28939--28956},
  year={2025},
  publisher={IEEE}
}

@article{Wan26,
  author={Wang, H. and Wang, N. and Li, X.},
  title={{FarmCD}: Agricultural Information-Guided Gated Network for Farmland Change Detection From Remote Sensing Images},
  journal={IEEE Transactions on Geoscience and Remote Sensing},
  year={2026},
  volume={61},
  pages={1--14},
  doi={10.1109/TGRS.2026.3652586}
}

@article{Yua26,
  author={Yuan, X. and Chen, L. and Zhang, J. and Zhou, G. and Wang, M. and Li, L.},
  title={IMEA-Net: An edge-sensitive network for cropland change detection in high-resolution remote sensing images},
  journal={ISPRS Journal of Photogrammetry and Remote Sensing},
  year={2026},
  volume={236},
  pages={175--196},
  doi={10.1016/j.isprsjprs.2026.03.044}
}

@article{Sui26,
  author={Sui, S. and Zhang, J. and Gu, H. and Chang, Y.},
  title={PANet: A multi-scale temporal decoupling network and its high-resolution benchmark dataset for detecting pseudo changes in cropland non-agriculturalization},
  journal={ISPRS Journal of Photogrammetry and Remote Sensing},
  year={2026},
  volume={233},
  pages={126--143},
  doi={10.1016/j.isprsjprs.2026.01.029}
}

@article{Li26,
  author={Li, G. and Han, P. and Wang, W. and Mu, T. and Xiao, Z. and Li, X.},
  title={HAM-CD: Hybrid Attention Mamba for Remote Sensing Change Detection},
  journal={IEEE Transactions on Geoscience and Remote Sensing},
  year={2026},
  volume={64},
  pages={5609518},
  doi={10.1109/TGRS.2026.3665418}
}

@article{Jia26,
  author={Jia, X. and Chen, Z. and Zhang, S. and Xue, X.},
  title={Change-LISA: Language-guided reasoning for remote sensing change detection},
  journal={IEEE Transactions on Geoscience and Remote Sensing},
  year={2026},
  volume={64},
  pages={3001415},
  doi={10.1109/TGRS.2026.3684817}
}

@article{Fang26,
  author={Fang, C. and Cheng, S. and Du, A. and Wu, C. and Ding, Y.},
  title={LGMM-Net: A local–global encoder and mask Mamba decoder network for remote sensing change detection},
  journal={IEEE Transactions on Geoscience and Remote Sensing},
  year={2026},
  volume={64},
  pages={2000923},
  doi={10.1109/TGRS.2026.3662322}
}

@article{Dong26,
  author={Dong, S. and Lu, C. and Fu, S. and Meng, X.},
  title={Synergy of Content and Style: Enhanced Remote Sensing Change Detection via Disentanglement and Refinement},
  journal={IEEE Transactions on Geoscience and Remote Sensing},
  year={2026},
  volume={64},
  pages={5610316},
  doi={10.1109/TGRS.2026.3664457}
}

@article{Cao26,
  author={Cao, Z. and Huang, Y. and Ma, L. and Zhou, Y. and Zhou, P. and Shi, W.},
  title={WFCDCLIP: A CLIP-Based Framework for Weakly Supervised Farmland Change Detection},
  journal={IEEE Transactions on Geoscience and Remote Sensing},
  year={2026},
  volume={64},
  pages={3684850},
  doi={10.1109/TGRS.2026.3684850}
}

@article{geminiteam2024gemini,
  author={{Gemini Team} and Anil, R. and Borgeaud, S. and Alayrac, J.-B. and Yu, J. and Soricut, R. and Schalkwyk, J. and Dai, A. M. and Hauth, A. and Millican, K. and Silver, D. and Johnson, M. and Antonoglou, I. and Schrittwieser, J. and Glaese, A. and Chen, J. and Pitler, E. and Lillicrap, T. and Lazaridou, A. and Firat, O. and others},
  title={Gemini: A Family of Highly Capable Multimodal Models},
  journal={arXiv preprint arXiv:2312.11805},
  year={2024},
  doi={10.48550/arXiv.2312.11805}
}
